\documentclass[5p,times,twocolumn]{elsarticle}

\biboptions{sort&compress}
\usepackage{amsmath,amsfonts,amssymb,bm}
\usepackage{algorithmic,algorithm}
\usepackage{array,graphicx,booktabs,makecell,multirow}
\usepackage{subfig}
\usepackage{placeins}
\usepackage{textcomp,url,verbatim,pifont,dsfont}
\usepackage{xltabular}
\keepXColumns
\usepackage{enumitem}
\usepackage[normalem]{ulem}
\useunder{\uline}{\ul}{}
\usepackage{xcolor}
\usepackage[hidelinks]{hyperref}

\newcommand{\figref}[1]{Figure~\ref{#1}}
\newcommand{\tabref}[1]{Table~\ref{#1}}

\newcommand{\revision}[1]{#1}
\newcolumntype{Y}{>{\centering\arraybackslash}p{1.25cm}}
\newcolumntype{C}{>{\centering\arraybackslash}m{1.6cm}}
\journal{Information Fusion}

\begin{document}
\begin{frontmatter}
\title{A Clinical Point Cloud Paradigm for In-Hospital Mortality Prediction from Multi-Level Incomplete Multimodal EHRs}

\author[buaa]{Bohao Li\corref{cor1}}
\ead{libh@buaa.edu.cn}
\cortext[cor1]{Corresponding author.}
\author[buaa]{Tao Zou}
\ead{zoutao@buaa.edu.cn}
\author[polyu]{Junchen Ye}
\ead{junchen.ye@polyu.edu.hk}
\author[buaa]{Yan Gong}
\ead{gongy@buaa.edu.cn}
\author[buaa]{Bowen Du}
\ead{dubowen@buaa.edu.cn}

\affiliation[buaa]{organization={Beihang University},city={Beijing},country={China}}
\affiliation[polyu]{organization={The Hong Kong Polytechnic University},city={Hong Kong},country={China}}

\begin{abstract}
Multimodal electronic health records (EHRs) support clinical risk prediction, but reliable information fusion is challenged by irregular sampling, missing modalities, and sparse outcome labels.
These coupled forms of incompleteness disrupt temporal alignment, restrict cross-modal interactions, and limit supervision.
Existing approaches typically address only a subset of these challenges, leaving unified event-level fusion under multi-level incompleteness insufficiently explored.
To uniformly model multi-level incomplete EHRs, we propose \textit{HealthPoint} (\textit{HP}), a novel unified Clinical Point Cloud Paradigm. Specifically, HP reformulates heterogeneous clinical events as independent points within a continuous \revision{quasi-4D} coordinate system spanned by content, time, modality, and case dimensions. 
To quantify interaction relationships between arbitrary point pairs within this coordinate system,
we introduce a \textit{Low-Rank Relational Attention} mechanism to efficiently couple high-order dependencies across the four dimensions. Then, a \textit{hierarchical interaction and sampling strategy} is used to balance the representation granularity of the point cloud with computational efficiency. Consequently, this paradigm supports flexible event-level interactions and fine-grained self-supervision, thereby naturally accommodating EHR heterogeneity, integrating multi-source information for robust modality recovery, and deeply utilizing unlabeled data. 
Extensive experiments on large-scale EHR datasets for risk prediction demonstrate that HP consistently achieves state-of-the-art performance and superior robustness under varying degrees of incompleteness.
\end{abstract}

\end{frontmatter}

\section{Introduction}
\label{section-1}

Electronic health records (EHRs) integrate physiological time series, clinical notes, and medical imaging, providing complementary information about patient status \cite{johnson2016mimic}. Reliable fusion of these heterogeneous sources is central to clinical risk prediction, particularly when observations are incomplete and temporally misaligned.
Multimodal pretraining has shown the value of jointly modeling clinical time series and notes for patient representation learning \cite{king2023multimodal}.

\begin{figure*}[!htbp]
    \centering
\includegraphics[width=\textwidth]{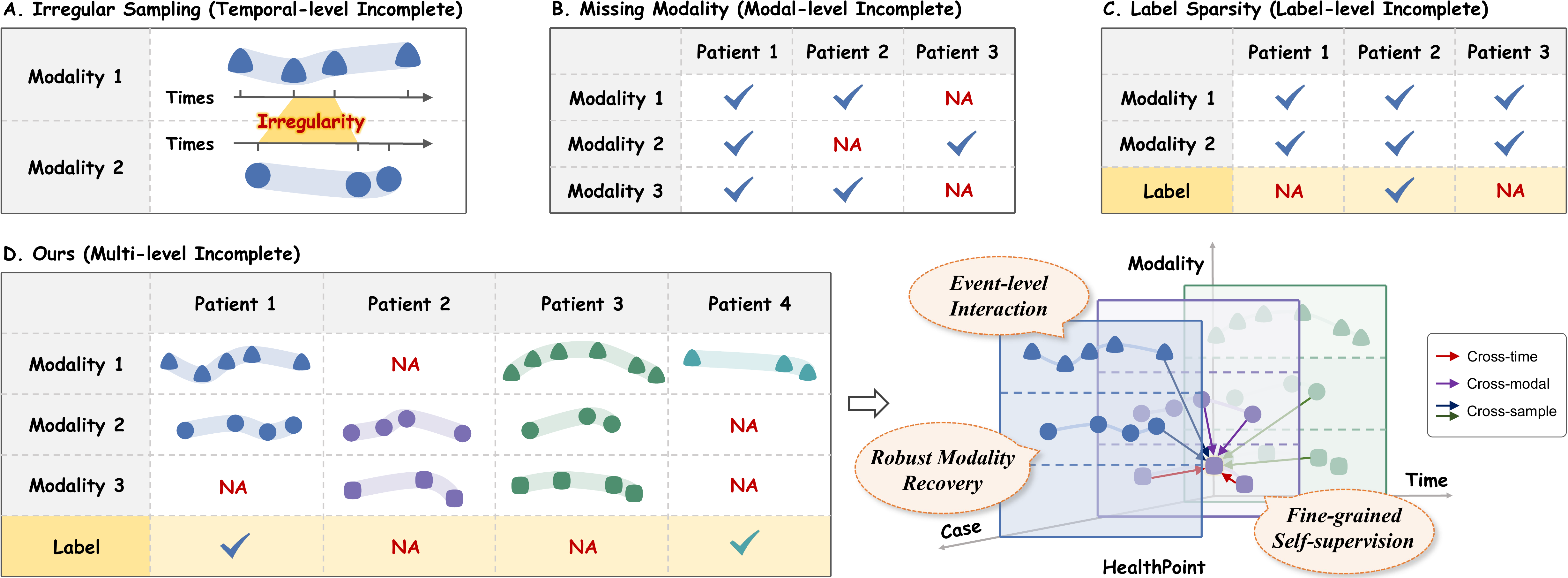}
    \caption{Irregular sampling, missing modality, and sparse label jointly result in multi-level incomplete multimodal clinical data. 
    HealthPoint addresses these challenges by modeling clinical events as a point cloud with learnable multi-dimensional relations, enabling event-level cross-domain interactions, robust modality recovery, and fine-grained self-supervision.
    }
    \label{fig:introduction}
\end{figure*}

\revision{However, \textit{real-world multimodal EHRs are pervasively incomplete} due to privacy regulations, device constraints, and diverse clinical workflows \cite{zhang2022m3care, zhang2023improving, li2025prime}. }
As shown in Figure~\ref{fig:introduction}(a–c), 
this incompleteness arises from three coupled factors: 
(1)~\textit{irregular sampling}, where clinical events are recorded at non-uniform intervals \cite{johnson2016mimic}; 
(2)~\textit{missing modality}, where the availability of different modalities varies across patients \cite{le2025multimodal}; and 
\revision{(3)~\textit{label sparsity}, where a large portion of records lack explicit diagnostic or outcome annotations \cite{wang2023hierarchical}. }
Together, these factors not only result in sparse and fragmented observations but also trigger \textit{cascading modeling failures}: including temporal distortion in disease evolution modeling \cite{zhang2023improving}, modal collapse during fusion \cite{zhang2022m3care}, and biased representations under scarce supervision \cite{li2025prime}, severely challenging risk prediction.

To address different forms of incompleteness, prior studies have explored several directions.
Specifically, irregular time-series modeling enhances robustness to non-uniform sampling \cite{zhang2023improving, che2018recurrent}. 
For modality missingness, some approaches reconstruct missing modalities using similar patient priors or observed modalities \cite{zhang2022m3care, wu2024multimodal, sun2024redcore, zhao2025diffmv}, while others adopt structured designs to ignore absent inputs \cite{yao2024drfuse, xu2024flexcare}. 
To mitigate label sparsity, self-supervised objectives, such as reconstruction or cross-modal alignment, are introduced as surrogate supervision signals \cite{zong2024self, li2025prime, wang2023hierarchical, xu2023vecocare}.


While prior strategies have shown promise, they typically address only one or two types of incompleteness \cite{lee2023learning, wu2024multimodal, li2025prime}. 
However, in real-world clinical practice, \textit{irregular sampling}, \textit{missing modality}, and \textit{label sparsity} pervasively co-occur, rendering approaches that require at least one form of completeness assumption incompatible with real-world EHR modeling requirements.
This motivates an event-level fusion framework that retains the timing and modality of observed clinical events while learning from partially observed patient records.
\textit{Accurate and robust mortality risk prediction under such multi-level incompleteness} remains an open and underexplored problem.

{\clubpenalty=0
 \widowpenalty=0
 \displaywidowpenalty=0
To address this problem, we identify the following three challenges:
(1)~\textit{\textbf{Heterogeneity induced by incompleteness.}} 
Multi-level incompleteness leads to inconsistent temporal patterns and modality combinations across patients, resulting in heterogeneous data structures without fixed topology. 
(2)~\textit{\textbf{Trade-off between modeling granularity and efficiency.}}
Accurate EHR modeling requires tracking continuous patient-state evolution, which necessitates fine-grained event-level representations beyond modality-level summarization \cite{shmatko2025learning, makarov2025large}. 
Yet, at this granularity, computational cost inevitably scales with the number of clinical events.
(3)~\textit{\textbf{Complexity of multi-relational modeling.}}
Multi-level incompleteness encourages exploiting cross-time, cross-modal, and even cross-patient consistency/similarity as surrogate constraints and multi-source fusion signals.
Yet, these dependencies are tightly coupled across time, modality, and patients, making unified representation non-trivial.
}

Intriguingly, we observe a structural resemblance between incomplete EHRs and 3D point clouds \cite{qi2017pointnet++}, as both form sparse sets without fixed topology. Motivated by the conceptual advantages of local relation modeling and neighborhood sampling in Point Transformers \cite{zhao2021point}, we propose \textbf{HealthPoint} (\textbf{HP}), a novel EHR-oriented paradigm for mortality risk prediction under multi-level incompleteness, which is fundamentally different from 3D point cloud modeling.

HP reconceptualizes each clinical event (observation) as a point in a unified \revision{\textit{quasi-4D} clinical coordinate system, where ``\textit{quasi}'' indicates four semantic rather than Euclidean dimensions: content, timestamp, modality, and patient case}.
To quantify dependencies between point pairs in this space, we introduce a \textit{Low-Rank Relational Attention} mechanism that approximates high-order interactions via compact multiplicative subspaces.
To balance granularity and efficiency, we further adopt a \textit{hierarchical interaction and sampling strategy} that adaptively focuses on salient events.
Built on this point-cloud framework with flexible event-level interactions, the paradigm naturally accommodates structural heterogeneity and supports fine-grained self-supervision and robust missing modality recovery, enabling effective learning from incomplete EHRs.
Experiments on two large-scale datasets support HP's predictive performance and robustness under the evaluated missing-data conditions.
Our contributions are summarized as follows.

\begin{itemize}[leftmargin=1.2em]
    \item \textit{\textbf{A clinical point cloud paradigm}} is proposed to address multi-level incompleteness in EHRs.
    By modeling clinical observations as points, HP enables flexible \textit{event-level interactions} that naturally handle irregular sampling and missing modality.
    On top of these interactions, we design \textit{fine-grained self-supervision} at the observation level, which facilitates \textit{robust modality recovery} and effective exploitation of unlabeled records.
    Through this tightly coupled design, HP \textit{simultaneously} addresses irregular sampling, missing modality, and label sparsity.

    \item \textit{\textbf{A low-rank relational attention mechanism}} is designed to quantify dependencies between arbitrary point pairs, thereby enabling event-level interactions in the clinical point space.
    By coupling multi-dimensional relative relations through a compact set of learnable feature vectors, this mechanism models high-order dependencies while keeping the interaction cost low.

    \item \textit{\textbf{A hierarchical interaction and sampling framework}} is introduced.
    Interactions are performed over hierarchical local clinical event neighborhoods, coupled with two learnable downsampling layers to extract representative clinical features. 
    This design enables effective patient's condition modeling while resolving the trade-off between granularity and efficiency.

    \item \textit{\textbf{A fine-grained self-supervised learning strategy}} is built upon the point cloud to address incompleteness.
    Observation-level objectives, including fine-grained alignment and reconstruction, exploit intrinsic self-constraints to leverage unlabeled data.
    Meanwhile, alignment mitigates cross-modality irregularity, while reconstruction supports robust missing-modality recovery.
    
\end{itemize}

\section{Preliminary}

Herein, we formulate the mortality risk prediction problem on multimodal EHRs with irregular sampling, missing modalities, and sparse labels.

\begin{figure*}[!htbp]
    \centering
\includegraphics[width=\textwidth]{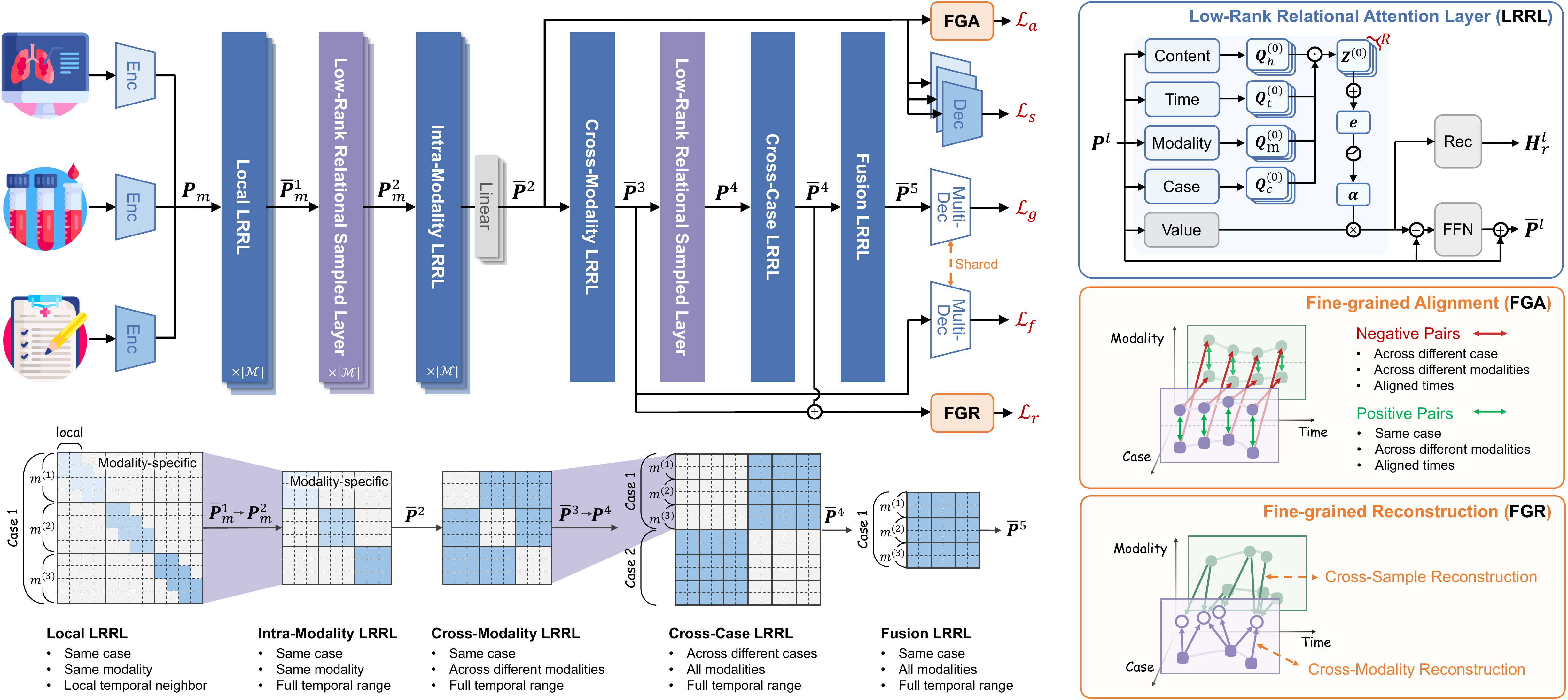}
    \caption{The framework of HP, where our code can be found in \url{https://anonymous.4open.science/r/HealthPoint}.}
    \label{fig:model}
\end{figure*}

\vspace{0.5em} 
\textbf{Clinical Event.}
We represent the EHR data as a set of discrete clinical events. 
Formally, each event is defined as a tuple $\mathtt{e}_k = (\bm{x}_k, t_k, \mathtt{m}_k, c_k)$, where $\bm{x}_k$ denotes the raw clinical content, $t_k \in \mathbb{R}$ is the timestamp, 
\revision{$\mathtt{m}_k \in \mathcal{M}=\{m^{(1)}, m^{(2)},\ldots \}$} indicates the modality type, 
and $c_k$ denotes the patient case to which the event belongs; hereafter, each ``case'' refers to one patient sample.
All events within a mini-batch are collected into $\mathcal{E} = \{ \mathtt{e}_k \}_{k=1}^{N}$.



\vspace{0.3em}
\textbf{Incompleteness \& Objective.}
For each case $c$, we introduce binary indicators $\mu_c^m \in \{0,1\}$ and $\ell_c \in \{0,1\}$, 
where $\mu_c^m=1$ indicates that modality $m$ is observed for case $c$, and $\ell_c=1$ indicates that the label $y_c$ is available.
Irregular sampling is reflected by the non-uniform timestamps $t_k$.
Given $\mathcal{E}$ with sparse availability $\{\bm{\mu}, \bm{\ell}\}$, our goal is to learn robust case-level representations for accurate risk prediction.

\section{Methodology}
\label{section-3}

We propose \textbf{HealthPoint} (HP), a unified framework that formulates incomplete multimodal EHR modeling as a clinical point cloud learning problem, as illustrated in Figure \ref{fig:model}.
HP embeds each clinical observation as a point in a \revision{quasi-4D} coordinate system defined over \revision{four semantically heterogeneous dimensions}: content, time, modality, and case.
To model high-order dependencies among arbitrary points in this space, we introduce \textit{Low-Rank Relational Attention}, which supports flexible event-level interactions.
Furthermore, a \textit{hierarchical interaction and sampling strategy} is employed to balance representation granularity with efficiency. 
Finally, we incorporate Fine-grained Alignment (FGA) and Reconstruction (FGR) objectives to effectively learn from incomplete data. 

\subsection{\textbf{Clinical Point Construction}}
\revision{We first map raw clinical event content $\bm{x}_k$ into modality-specific feature representations $\bm{h}_k \in \mathbb{R}^{d_{\mathtt{m}_k}^{enc}}$, where $d_{\mathtt{m}_k}^{enc}$ denotes the embedding dimension of modality $\mathtt{m}_k$. Specifically, we use a two-layer MLP for vital signs and lab tests, Clinical-Longformer~\cite{li2022clinical} for clinical notes, and DenseNet~\cite{cohen2020limits} for medical imaging.
}
Consequently, we obtain the event token set $\bm{H} = \{\bm{h}_k\}_{k=1}^{N}$.

Then, each event $\mathtt{e}_k$ is conceptualized as a \textbf{clinical point} by assigning its representation $\bm{h}_k$ a unique coordinate tuple:
\begin{equation}
    p_k = (\bm{h}_k, t_k, \mathtt{m}_k, c_k),
\end{equation}
within the clinical point cloud space.
Here, $\bm{h}_k$ serves as the content (feature) coordinate, while $t_k, \mathtt{m}_k, c_k$ denote the temporal, modal, and case coordinates, respectively. 
Accordingly, the global token set $\bm{H}$ corresponds to a coordinate set $\bm{P} = \{ p_k \}_{k=1}^N$.


\revision{For notational convenience, we use $\bm{H}_{m}^c$ and $\bm{P}_{m}^c$ to denote the token and coordinate sequences of case $c$ under modality $m$, and $\bm{H}_{m}$ and $\bm{P}_{m}$ to denote the collections of modality $m$.}

\subsection{\textbf{Low-Rank Relational Attention Layer}}
\label{sec:lrrl}
To enable event-level interactions in this space, we propose the \textit{Low-Rank Relational Attention Layer} (\textbf{LRRL}) as the core component of HP, which quantifies pairwise relations between points.
Formally, the $l$-th layer operates as:
{ 
\setlength{\abovedisplayskip}{3pt} 
\setlength{\belowdisplayskip}{3pt}
\begin{equation}
    (\bar{\bm{H}}^l, \bar{\bm{P}}^l) = \operatorname{LRRL}^l(\bm{H}^l, \bm{P}^l),
\end{equation}
}
where $\bm{H}^l,\bm{P}^l$ are the input token and coordinate sets, $\bar{\bm{H}}^l,\bar{\bm{P}}^l$ are the outputs, and only the content $\bm{h}$ within $\bm{P}^l$ is updated.
\revision{
Unlike spatial points whose neighborhoods and interaction relations can be defined by Euclidean distances~\cite{zhao2021point}, clinical points involve heterogeneous \textit{content}, \textit{time}, \textit{modality}, and \textit{case} relations, making their interaction nontrivial to define.
To address this, HP adopts progressive point-wise interactions with predefined layer-specific neighborhoods. 
For the $l$-th LRRL, we use $\mathcal{N}^l(i)$ to denote the neighborhood of the center point $p_i$.
Within each neighborhood, LRRL first extracts per-dimension relational features and then fuses them via low-rank coupling to approximate high-order pairwise relations.}

\vspace{0.3em}
\noindent \textbf{Multi-dimensional Relational Features.}
\revision{
For any point pair $(p_i,p_j)$ with $j \in \mathcal{N}^l(i)$ and content $\bm{h}_i,\bm{h}_j \in \mathbb{R}^{d^l}$, where $d^l$ denotes the feature dimension of the $l$-th layer, we define relational features $\bm{r}_{ij}^{*} \in \mathbb{R}^{d^l}$ for the four candidate dimensions:}

\begin{itemize}[leftmargin=1.2em, topsep=1pt]
    \item \textbf{Content ($h$):} Captures clinical content relations via query-key interaction, formulated as $\bm{r}_{ij}^{h} = \mathbf{W}_Q \bm{h}_i - \mathbf{W}_K \bm{h}_j$~\cite{zhao2021point}, \revision{where $\mathbf{W}_Q,\mathbf{W}_K \in \mathbb{R}^{d^l \times d^l}$ are layer-specific projections}.
    
    \item \textbf{Time ($t$):} Evaluates the time interval $\Delta t_{ij} = t_i - t_j$, encoded by a two-layer MLP $\phi_t$ as $\bm{r}_{ij}^{t} = \phi_t(\Delta t_{ij})$ \cite{zhang2023warpformer},
    \revision{where $\phi_t$ maps the time interval to $\mathbb{R}^{d^l}$.}
    
    \item \textbf{Modality ($\mathtt{m}$):} Learns modality relationships by querying a learnable affinity matrix \revision{$\mathbf{E}_{\mathtt{m}} \in \mathbb{R}^{|\mathcal{M}|\times|\mathcal{M}| \times d^l}$}, denoted as $\bm{r}_{ij}^{\mathtt{m}} = \mathbf{E}_{\mathtt{m}}[\mathtt{m}_i, \mathtt{m}_j]$.
    

    \item \textbf{Case ($c$):} Quantifies case similarity based on disease evolution patterns.
    For a case pair $(c_i, c_j)$, the relation embedding is computed by: $\bm{r}_{ij}^{c} = \frac{1}{|\mathcal{V}_{ij}|} \sum_{m \in \mathcal{V}_{ij}} \text{BiGRU} ( \bm{H}_{m}^{c_i} - \bm{H}_{m}^{c_j} )$,
    where $\mathcal{V}_{ij} = \{ m \mid \mu_{c_i}^m \cdot \mu_{c_j}^m = 1, m\in \mathcal{M} \}$ denotes the set of co-observed modalities.
    \revision{And, the neighborhood definition ensures $\mathcal{V}_{ij}\neq\emptyset$.}
    Here, $\bm{H}_{m}^{c_i}$ and $\bm{H}_{m}^{c_j}$ are temporally aligned event sequences (via the sampling operation in the subsequent LRRSL), and their difference reflects trajectory deviation, encoded by a BiGRU \cite{cho2014learning}, \revision{where the BiGRU hidden/output dimension is set to $d^l$.} 
\end{itemize}
\revision{
Depending on the layer-specific neighborhood $\mathcal{N}^l(i)$, some relational dimensions are naturally inactive and therefore not computed; for example, the case dimension is excluded when cross-case interaction is absent. Formally, $\mathcal{D}^l \subseteq \{h,t,\mathtt{m},c\}$ denotes the active relational dimensions of the $l$-th layer, and only relations $\bm{r}_{ij}^{*}$ with $* \in \mathcal{D}^l$ are activated. The concrete definitions of $\mathcal{N}^l(i)$ and $\mathcal{D}^l$ are provided in Section~\ref{sec:hierarchical}.}

\vspace{0.5em}
\noindent \textbf{Low-Rank Coupling.}
To couple these \revision{active relational features $\{\bm{r}_{ij}^{*}\}_{* \in \mathcal{D}^l}$} into a unified attention logit,
we adopt the Canonical Polyadic (CP) decomposition~\cite{kolda2009tensor} to perform a $R$-rank approximation of this underlying high-order interaction tensor.
For each rank $\gamma \in \{1, \dots, R\}$ and dimension $* \in \mathcal{D}^l$, we introduce learnable projection vectors $\mathbf{Q}_{*}^{(\gamma)} \in \mathbb{R}^{d^l}$.
Then, the joint attention logit $e_{ij}$ is computed by aggregating the coupled products across all ranks:
\begin{align}
    Z_{ij}^{(\gamma )} &= \prod\nolimits_{* \in \mathcal{D}^l} \langle \mathbf{Q}_{*}^{(\gamma )}, \bm{r}_{ij}^{*} \rangle, \\
    e_{ij} &= \sum\nolimits_{\gamma =1}^{R} Z_{ij}^{(\gamma )} + \sum\nolimits_{* \in \mathcal{D}^l} \mathbf{w}_{*}^\top \bm{r}_{ij}^{*} + b
\label{eq:low_rank}
\end{align}
where $\langle \cdot, \cdot \rangle$ denotes the dot product. 
The \textit{coupled term} $\sum_{\gamma =1}^{R} Z_{ij}^{(\gamma)}$ represents the relational coefficient aggregated from $R$ latent factors, fusing multi-dimensional dependencies non-linearly. 
\revision{Complementarily, the \textit{unary term} $\mathbf{w}_{*}^\top \bm{r}_{ij}^{*}$ captures the first-order contribution of each active dimension, where $\mathbf{w}_{*}\in\mathbb{R}^{d^l}$ is a learnable weight vector and $b\in\mathbb{R}$ is a learnable scalar bias.}
Additionally, by adjusting the dimensions of $\bm{r}_{ij}^{*}$, this attention can be easily extended to a multi-head version.
Finally, point features are updated via attention aggregation followed by \revision{a two-layer MLP}:
\begin{align}
    \alpha_{ij} &= \operatorname*{Softmax}_{j \in \mathcal{N}^l(i)}(e_{ij}), \\
    \quad \bar{\bm{h}}^{l}_i &= \text{MLP} [ \bm{h}^{l}_i + \sum\nolimits_{j \in \mathcal{N}^l(i)} \alpha_{ij} (\mathbf{W}_V \bm{h}^{l}_j) ],
\end{align}
where $\bar{\bm{h}}^{l}_i \in \bar{\bm{H}}^{l}$,
\revision{and $\mathbf{W}_V \in \mathbb{R}^{d^l \times d^l}$ is the layer-specific value projection; layer/head superscripts are omitted for readability.}

\subsection{\textbf{Hierarchical Interaction and Sampling}}
\label{sec:hierarchical}
To circumvent the prohibitive cost of global interactions while capturing multi-granularity, temporally aligned disease dynamics, we propose a hierarchical framework with a learnable sampling mechanism and a five-level interaction strategy.

\vspace{0.5em}
\noindent \textbf{Low-Rank Relational Sampled Layer (LRRSL).}
To control the granularity of clinical token sequences and balance computational costs, we introduce \textbf{LRRSL} to compress the point token sequence, drawing inspiration from 3D point cloud sampling \cite{zhao2021point}. 
Formally, the LRRSL operation after the $l$-th LRRL is defined as:
\begin{equation}
    (\bm{H}^{(l+1)}, \bm{P}^{(l+1)}) = \operatorname{LRRSL}^l(\bar{\bm{H}}^l, \bar{\bm{P}}^l, \mathcal{A}^l)
\end{equation}
where $\mathcal{A}^l$ is a virtual point set serving as sampling anchors.

Due to the consistency of the sampling mechanism across modalities and cases, we exemplify the process using the token subset $\bm{H}_{m}^c \subset \bar{\bm{H}}^l$ and its corresponding anchor subset $\mathcal{A}^l_{m} \subset \mathcal{A}^l$.
Each anchor $a_i \in \mathcal{A}^l_{m}$ is defined as $a_i = (t_i, \bm{q}_{m}^l)$, where $t_i$ is drawn from a fixed temporal grid $\mathcal{T}^l = \{0, \Delta t^l_{m}, 2\Delta t^l_{m}, \dots\}$, and $\bm{q}_{m}^l \in \mathbb{R}^{d^l}$ is a learnable modality-specific query in the current LRRSL, \revision{where $d^l$ is the  dimension of the current layer}.

For a specific anchor $a_i = (t_i, \bm{q}_{m}^l)$ and a clinical point token $\bm{h}_j \in \bm{H}_{m}^c$ (with coordinate $p_j$), the sampling interaction relies solely on the content and time dimensions:
\begin{itemize}[leftmargin=1.2em, topsep=1pt]
    \item \textbf{Content:} Captures key content via $\bm{r}_{ij}^{h} = \mathbf{W}_Q \bm{q}_{m}^l - \mathbf{W}_K \bm{h}_j$. 

    \item \textbf{Time:} Measures temporal proximity via $\bm{r}_{ij}^{t} = \phi_t(t_{i} - t_j)$.
\end{itemize}

Then, similar to LRRL, the sampling process is given:
\begin{align}
e_{ij}
&= \sum_{\gamma=1}^{R}
\prod_{* \in \{h,t\}}
\langle \mathbf{Q}_{*}^{(\gamma)}, \bm{r}_{ij}^{*} \rangle
 + \sum_{* \in \{h,t\}} \mathbf{w}_{*}^{\top}\bm{r}_{ij}^{*} + b, \\
\bm{h}_{i}^{(l+1)}
&= \sum\nolimits_{\bm{h}_j \in \bm{H}_{m}^{c}}
\operatorname{Softmax}_{j}(e_{ij})
\bigl(\mathbf{W}_{V}\bm{h}_{j}\bigr).
\end{align}
\revision{Here, $\mathbf{W}_Q,\mathbf{W}_K,\mathbf{W}_V \in \mathbb{R}^{d^l \times d^l}$ are LRRSL-specific projections, and $\phi_t$ maps time intervals to $\mathbb{R}^{d^l}$; layer/head indices are also omitted for readability.}

Consequently, for case $c$ and modality $m$ at anchor position $a_i$, we obtain a sampled token $\bm{h}_{i}^{(l+1)} \in \bm{H}^{(l+1)}$. This forms a new coordinate tuple: ${p}_{i}^{(l+1)} = (\bm{h}_{i}^{(l+1)}, t_i, m, c) \in \bm{P}^{(l+1)}$. 
These sampled points capture the temporal evolution of the condition, offering a controllable density via the interval $\Delta t^l_{m}$.


\vspace{0.5em}
\noindent \textbf{Hierarchical Interaction.}
To facilitate progressive interactions and further mitigate costs, we design a five-level hierarchical interaction strategy. 
Our structure follows the fundamental principle of prioritizing intra-modality aggregation before cross-modality fusion \cite{baltruvsaitis2018multimodal}. 
Specifically, building upon LRRL and LRRSL, we instantiate the HP backbone. \revision{The first two layers are modality-specific; for compactness, $\operatorname{LRRL}_{m}^l$ and $\operatorname{LRRSL}_{m}^l$ denote the modules for modality $m$ at layer $l$, with operating feature dimension $d_m^l$.} We define each layer's computation, interaction neighborhood $\mathcal{N}^l(i)$, and active dimensions $\mathcal{D}^l$ as follows:
\begin{itemize}[leftmargin=1.2em, topsep=1pt]
    \item \textbf{Local LRRL.} Captures fine-grained short-term consistency within a time window $\delta$. Here, $\mathcal{N}^1(i) = \{j \mid c_j=c_i, \mathtt{m}_j=\mathtt{m}_i, |t_i - t_j| \le \delta\}$ and $\mathcal{D}^1 = \{h, t\}$.
    \revision{This layer is modality-specific and executes:
    $( \bar{\bm{H}}_{m}^1, \bar{\bm{P}}_{m}^1 )
    = \operatorname{LRRL}_{m}^1(\bm{H}_{m}, \bm{P}_{m})$,
    followed by
    $( \bm{H}_{m}^2, \bm{P}_{m}^2 )
    = \operatorname{LRRSL}_{m}^1(\bar{\bm{H}}_{m}^1, \bar{\bm{P}}_{m}^1, \mathcal{A}_{m}^1)$.}
    
    \item \textbf{Intra-Modality LRRL.} Models long-term dependencies within specific modalities, defined by $\mathcal{N}^2(i) = \{j \mid c_j=c_i, \mathtt{m}_j=\mathtt{m}_i\}$ and $\mathcal{D}^2 = \{h, t\}$. 
    \revision{
    The operation is also modality-specific:
    $( \bar{\bm{H}}_{m}^2, \bar{\bm{P}}_{m}^2 )
    = \operatorname{LRRL}_{m}^2(\bm{H}_{m}^2, \bm{P}_{m}^2)$.
    Then, a linear projection maps modality-specific features into a unified dimension $d^u$, and the outputs are collected into $(\bar{\bm{H}}^2,\bar{\bm{P}}^2)$ for subsequent cross-modality interaction.}
    
    \item \textbf{Cross-Modality LRRL.} Fuses complementary multi-modal information, with $\mathcal{N}^3(i) = \{j \mid c_j=c_i, \mathtt{m}_j \neq \mathtt{m}_i\}$ and $\mathcal{D}^3 = \{h, t, \mathtt{m}\}$. 
    The process involves $( \bar{\bm{H}}^3, \bar{\bm{P}}^3 ) = \operatorname{LRRL}^3(\bar{\bm{H}}^2, \bar{\bm{P}}^2)$, followed by $( \bm{H}^4, \bm{P}^4 ) = \operatorname{LRRSL}^3(\bar{\bm{H}}^3, \bar{\bm{P}}^3, \mathcal{A}^3)$.
    
    \item \textbf{Cross-Case LRRL.} Retrieves latent priors from similar patients, where \revision{$\mathcal{N}^4(i) = \{j \mid c_j \neq c_i, \mathcal{V}_{ij}\neq\emptyset \}$} and $\mathcal{D}^4 = \{h, t, \mathtt{m}, c\}$. 
    This is formulated as $( \bar{\bm{H}}^4, \bar{\bm{P}}^4 ) = \operatorname{LRRL}^4(\bm{H}^4, \bm{P}^4)$.
    
    \item \textbf{Fusion LRRL.} Performs global aggregation for the final representation, with $\mathcal{N}^5(i) = \{j \mid c_j=c_i\}$ and $\mathcal{D}^5 = \{h, t, \mathtt{m}\}$. 
    The final output is derived via $( \bar{\bm{H}}^5, \bar{\bm{P}}^5 ) = \operatorname{LRRL}^5(\bar{\bm{H}}^4, \bar{\bm{P}}^4)$.
\end{itemize}

HP sequentially executes these layers to yield robust representations.  
\revision{
Notably, the first two layers are \textit{modality-specific} with $d^1_m=d^2_m=d_{m}^{enc}$, while layers 3--5 use a unified dimension after linear projection, i.e., $d^3=d^4=d^5=d^u$.}

\subsection{\textbf{Fine-grained Self-supervised Learning}}
Based on the point cloud paradigm, we obtain observation-level representations of patient dynamics, upon which self-supervised objectives are constructed.
This strategy fully exploits intrinsic constraints within incomplete EHR mini-batches to maximize the utilization of unlabeled data and alleviate modality missingness.

\vspace{0.5em}
\noindent \textbf{Fine-grained Alignment (FGA).}
To leverage unlabeled samples/cases, we introduce a fine-grained alignment objective that aligns disease evolution across modalities.
Crucially, this operates on the Intra-Modality LRRL output $\bar{\bm{H}}^2$ to prevent information leakage from subsequent cross-modal fusion.
The alignment loss $\mathcal{L}_{a}$ is formulated using a contrastive learning objective \cite{chen2020simple, li2025prime}:
\begin{equation}
    \resizebox{1.0\linewidth}{!}{$ 
        \displaystyle 
        \mathcal{L}_{a} = - \frac{1}{|\bar{\bm{H}}^2|} \sum_{\bm{h}_i \in \bar{\bm{H}}^2} \log \frac{\sum_{j \in \mathcal{P}^{+}(i)} e^{\sigma(\bm{h}_i, \bm{h}_j)/\tau}}{\sum_{j \in \mathcal{P}^{+}(i)} e^{\sigma(\bm{h}_i, \bm{h}_j)/\tau} + \sum_{n \in \mathcal{P}^{-}(i)} e^{\sigma(\bm{h}_i, \bm{h}_n)/\tau}}
    $}
\end{equation}
where $\bm{h}_i$ represents a valid clinical point within $\bm{H}^2$ (associated with patient $c_i$, modality $\mathtt{m}_i$, and timestamp $t_i$, subject to $\mu_{c_i}^{\mathtt{m}_i}=1$), $\sigma(\bm{u}, \bm{v}) = \frac{\bm{u}^\top \bm{v}}{\|\bm{u}\|\|\bm{v}\|}$ denotes the cosine similarity, and \revision{$\tau=0.1$ is the temperature parameter used to scale the cosine similarities}.
The positive set $\mathcal{P}^{+}(i)$ and negative set $\mathcal{P}^{-}(i)$ are strictly defined based on the unified coordinates:
\begin{itemize}[leftmargin=1.2em]
    \item \textbf{Positive Pairs $\mathcal{P}^{+}(i)$:} Points indexed by $j$ from the \textit{same case} ($c_j = c_i$) but \textit{different modalities} ($\mathtt{m}_j \neq \mathtt{m}_i$) at \textit{aligned times} ($t_j = t_i$), capturing shared underlying pathology.
    \item \textbf{Negative Pairs $\mathcal{P}^{-}(i)$:} Points indexed by $n$ from \textit{different cases} ($c_n \neq c_i$) and \textit{different modalities} ($\mathtt{m}_n \neq \mathtt{m}_i$) at \textit{aligned times} ($t_n = t_i$), serving as background negatives.
\end{itemize}

\vspace{0.5em}
\noindent \textbf{Fine-grained Reconstruction (FGR).}
To recover missing modalities, thereby preventing modal collapse and further mining cross-view constraints from unlabeled data, we propose the \textit{Fine-grained Reconstruction} objective. 
This mechanism reconstructs fine-grained evolutionary representations by leveraging Cross-Modality (Layer 3) and Cross-Case (Layer 4) interactions. 
Specifically, to decouple reconstruction from the primary update, we modify the LRRL architecture (Figure 2) by introducing a dedicated MLP, denoted as $\operatorname{REC}(\cdot)$, which operates on attention logits parallel to the standard path.
The reconstruction output $\bm{h}^{l}_r$ for layer $l \in \{3, 4\}$ is given as:
\begin{equation}
    \bm{h}^{l}_r = \text{REC} \left[ \sum\nolimits_{j \in \mathcal{N}^l(i)} \alpha_{ij} (\mathbf{W}_V \bm{h}^{l}_j) \right]
\end{equation}
yielding the reconstruction feature sets $\bm{H}^{3}_r$ and $\bm{H}^{4}_r$. 
Subsequently, we aggregate these multi-view recovery signals to form the complete reconstruction representation:
\begin{equation}
    \hat{\bm{H}} = \tilde{\bm{H}}^{3}_r + \bm{H}^{4}_r
\end{equation}
where $\tilde{\bm{H}}^{3}_r$, obtained via $( \tilde{\bm{H}}^{3}_r, \_ ) = \operatorname{LRRSL}^3(\bm{H}^{3}_r, \bar{\bm{P}}^3, \mathcal{A}^3)$, is downsampled to match the granularity of $\bm{H}^{4}_r$.
Finally, for valid modalities, we minimize the distance between $\hat{\bm{H}}$ and the Layer 4 output $\bar{\bm{H}}^4$, forcing the model to  infer missing information from cross-modal and cross-case contexts:
\begin{equation}
    \mathcal{L}_{r} = \sum_{c, m} \mu_c^m \cdot \| \hat{\bm{H}}_{m}^c - \bar{\bm{H}}_{m}^{4,c} \|_2^2,
\end{equation}
where $\hat{\bm{H}}_{m}^c \subset \hat{\bm{H}}$ and $\bar{\bm{H}}_{m}^{4,c} \subset \bar{\bm{H}}^4$.
For missing modalities, we update $\bar{\bm{H}}^4$ using $\hat{\bm{H}}$: $\bar{\bm{H}}^4 \leftarrow \bar{\bm{H}}^4 \odot \bm{\mu} + \hat{\bm{H}} \odot (1 - \bm{\mu})$, where $\odot$ denotes element-wise multiplication and $\bm{\mu}$ is the modality availability mask.




\subsection{\textbf{Optimization and Inference}}
\noindent \textbf{Supervised Objectives.} 
To ensure discriminative representations, we design multi-level supervision for labeled cases ($\ell_c=1$).
First, let $\bar{\bm{h}}^{l, c}_{m, last}$ denote the last-timestamp feature of the sequence $\bar{\bm{H}}^{l, c}_{m} \subset \bar{\bm{H}}^{l}$, and $\mathbf{u}^l_c = \operatorname{Concat}_{m} [ \bar{\bm{h}}^{l, c}_{m, last} ]$ be the fused representation.
We employ a \textit{shared classifier} $f_{\phi}$ for fusion layers and distinct \textit{modality-specific heads} $\{f_{m}\}$ for uni-modal branches.
The task loss is designed to capture information at different abstraction levels:

(1) \textbf{Global Fusion ($\mathcal{L}_g$):} Applied to Layer 5, this supervises the final representation \textit{enriched with cross-case priors} to ensure robust global reasoning: $\mathcal{L}_g = \sum_{c} \ell_c \cdot \operatorname{CE}(f_{\phi}(\mathbf{u}^5_c), y_c)$.

(2) \textbf{Cross-modal Fusion ($\mathcal{L}_f$):} Applied to Layer 3, this focuses on \textit{intra-case multi-modal} fusion, and the loss is formulated as: $\mathcal{L}_f = \sum_{c} \ell_c \cdot \mu_c^{all} \cdot \operatorname{CE}(f_{\phi}(\mathbf{u}^3_c), y_c)$, where we strictly require complete modality availability, defined as $\mu_c^{all} \triangleq \prod_{m\in\mathcal{M}} \mathds{1}_{\mu_c^m}$.

(3) \textbf{Uni-modal Regularization ($\mathcal{L}_s$):} To \textit{prevent modality collapse} where the model over-relies on dominant modalities, we force each modality to learn independent semantics on Layer 2 using sequence averaging: $\mathcal{L}_s = \sum_{c, m} \ell_c \cdot \mu_c^m \cdot \operatorname{CE}(f_{m}(\operatorname{Mean}(\bar{\bm{H}}^{2, c}_{m})), y_c)$.

The total loss function is given as follows:
\begin{equation}
    \mathcal{L}_{total} = (\mathcal{L}_g + \mathcal{L}_f + \mathcal{L}_s) + \lambda_a \mathcal{L}_{a} + \lambda_r \mathcal{L}_{r}, 
\end{equation}
where $\lambda_a$ and $\lambda_r$ are used to balance the self-supervised terms.

\vspace{0.5em}
\noindent \textbf{Adaptive Entropy-based Inference.} 
During the inference phase, we employ an adaptive selection strategy based on prediction confidence. 
We compute the entropy of predictions from all branches (Uni-modal, Cross-modal, and Global) \cite{shannon1948mathematical, devries2018learning}. 
The final prediction is selected as the one with the \textit{lowest entropy}, yielding the most confident output while mitigating potentially noisy imputations.

\section{Experiments}
\label{section-4}
We empirically evaluate \textbf{HP} under diverse incomplete EHR conditions, demonstrating its effectiveness over recent baselines.
In addition, we present ablations, a case study, and complexity analyses to further examine our method.

\subsection{\textbf{Experimental Settings}}

This section outlines our experimental settings, including the datasets, evaluation protocols, baseline methods, and implementation details.

\vspace{0.2em}
\textit{\textbf{Datasets.}}
We evaluate on two widely used large-scale EHR datasets: MIMIC-III \cite{johnson2016mimic} and MIMIC-IV \cite{johnson2023mimic}.
MIMIC-III provides physiological time series ($m^{(1)}$) and sequential clinical notes ($m^{(2)}$), while MIMIC-IV incorporates physiological signals ($m^{(1)}$), a discharge summary ($m^{(2)}$), and chest X-rays ($m^{(3)}$).
We follow standard preprocessing pipelines \cite{harutyunyan2019multitask, zhang2023improving, lee2023learning} to construct in-hospital mortality (IHM) prediction datasets with \revision{irregular sampling and inherent modality missingness}.
MIMIC-III uses observations from the first 48 hours after admission. For MIMIC-IV, the prediction landmark is discharge, and chest X-rays are selected from the preceding 48 hours; this constitutes a retrospective discharge-aligned evaluation.
To simulate label sparsity, we randomly drop 50\% of outcome labels. 
Dataset splits are 25,172/6,293/5,556 (MIMIC-III) and 22,033/5,445/3,408 (MIMIC-IV) for train/val/test.
The inherent modality missing rate in the MIMIC-III training set is reported as 56\%, rounded from $14{,}214/25{,}172\approx56.47\%$.
See \ref{appendix:datasets} for more details.

\vspace{0.2em}
\textit{\textbf{Evaluation Protocol.}}
We conduct binary classification for IHM prediction, reporting AUROC, AUPRC, and F1-score as evaluation metrics, following prior works \cite{zhang2023improving, king2023multimodal, li2025prime}.
To comprehensively evaluate performance under different incompleteness settings, we additionally construct variants on MIMIC-III by simulating:
(1)~varying label missing rates (25\%/50\%/75\%/90\%);
(2)~varying modality missing rates (56\%/75\%/90\%);
(3)~only modality missing; and
(4)~only label missing.
These setups are summarized in Table~\ref{tab:incomplete-settings}.

\setlength{\abovecaptionskip}{4pt} 

\begin{table}[b]
\centering
\caption{Incompleteness settings on MIMIC-III.}
\label{tab:incomplete-settings}
\resizebox{\columnwidth}{!}{
\begin{tabular}{lcc}
\toprule
\textbf{Setting} & \textbf{Label Missing} & \textbf{Modality Missing} \\
\midrule
Raw Dataset & 0\% & 56\% \\
Main Experiment & 50\% & 56\% \\
Varying label missing rate & 25\% / 50\% / 75\% / 90\% & 56\% \\
Varying modality missing rate & 50\% & 56\% / 75\% / 90\% \\
\revision{Only Modality Missing} & 0\% & 56\% \\
\revision{Only Label Missing} & 90\% & 0\% \\
\bottomrule
\end{tabular}
}
\end{table}

\begin{table*}[t]
\centering
\caption{\revision{Main results under incomplete settings on MIMIC-III and MIMIC-IV datasets.}}
\label{tab:main-results}
\resizebox{\textwidth}{!}{
\begin{tabular}{c|ccc|ccc|ccc}
\toprule
\multirow{2}{*}{\textbf{Method}} & \multirow{2}{*}{\makecell{\textbf{Irregular}}} & \multirow{2}{*}{\makecell{\textbf{Missing}\\\textbf{Modality}}} & \multirow{2}{*}{\makecell{\textbf{Missing}\\\textbf{Label}}} & \multicolumn{3}{c|}{\textbf{MIMIC-III}} & \multicolumn{3}{c}{\textbf{MIMIC-IV}} \\
 & & & & AUROC & AUPRC & F1 & AUROC & AUPRC & F1 \\
\midrule
MIPM~\cite{zhang2023improving} & \checkmark & & & $91.621_{\pm 0.041}$ & $67.197_{\pm 0.252}$ & $60.239_{\pm 0.236}$ & $97.693_{\pm 0.151}$ & $92.419_{\pm 0.218}$ & $86.501_{\pm 0.512}$ \\
PRIME~\cite{li2025prime} & \checkmark & & \checkmark & $91.537_{\pm 0.036}$ & $66.625_{\pm 0.394}$ & $59.518_{\pm 0.329}$ & $97.717_{\pm 0.040}$ & $92.338_{\pm 0.172}$ & $85.975_{\pm 0.395}$ \\
MEDHMP~\cite{wang2023hierarchical} & & & \checkmark & $90.091_{\pm 0.081}$ & $63.842_{\pm 0.603}$ & $55.423_{\pm 0.522}$ & $97.633_{\pm 0.035}$ & $91.873_{\pm 0.192}$ & $86.052_{\pm 0.506}$ \\
VecoCare~\cite{xu2023vecocare} & & & \checkmark & $90.234_{\pm 0.063}$ & $61.692_{\pm 0.457}$ & $55.522_{\pm 0.383}$ & $97.613_{\pm 0.057}$ & $92.386_{\pm 0.311}$ & $86.557_{\pm 0.481}$ \\
HEART~\cite{huang2024heart} & & \checkmark & & $90.222_{\pm 0.057}$ & $62.889_{\pm 0.371}$ & $56.893_{\pm 0.228}$ & $96.865_{\pm 0.063}$ & $91.639_{\pm 0.102}$ & $86.689_{\pm 0.217}$ \\
MuIT-EHR~\cite{chan2024multi} & & \checkmark & & $90.296_{\pm 0.059}$ & $62.957_{\pm 0.510}$ & $56.245_{\pm 0.441}$ & $96.918_{\pm 0.116}$ & $91.471_{\pm 0.304}$ & $85.961_{\pm 0.365}$ \\
M3Care~\cite{zhang2022m3care} & & \checkmark & & $90.357_{\pm 0.093}$ & $63.433_{\pm 0.388}$ & $57.201_{\pm 0.511}$ & $96.977_{\pm 0.105}$ & $91.597_{\pm 0.281}$ & $86.498_{\pm 0.305}$ \\
UMM~\cite{lee2023learning} & \checkmark & \checkmark & & $88.359_{\pm 0.064}$ & $59.492_{\pm 0.679}$ & $54.434_{\pm 0.653}$ & $97.323_{\pm 0.115}$ & $92.125_{\pm 0.322}$ & \uline{86.853}$_{\pm 0.671}$ \\
DrFuse~\cite{yao2024drfuse} & & \checkmark & & $89.819_{\pm 0.169}$ & $62.713_{\pm 0.859}$ & $57.359_{\pm 0.359}$ & $97.030_{\pm 0.021}$ & $91.292_{\pm 0.179}$ & $85.945_{\pm 0.309}$ \\
RedCore~\cite{sun2024redcore} & & \checkmark & & \uline{91.710}$_{\pm 0.069}$ & $67.169_{\pm 0.455}$ & \uline{60.316}$_{\pm 0.377}$ & \uline{97.816}$_{\pm 0.030}$ & $92.659_{\pm 0.123}$ & 86.547$_{\pm 0.331}$ \\
FlexCare~\cite{xu2024flexcare} & & \checkmark & & $91.637_{\pm 0.048}$ & \uline{67.242}$_{\pm 0.281}$ & $60.198_{\pm 0.218}$ & $97.013_{\pm 0.035}$ & $92.073_{\pm 0.089}$ & $86.430_{\pm 0.153}$ \\
Diffmv~\cite{zhao2025diffmv} & & \checkmark & & $91.464_{\pm 0.056}$ & $66.389_{\pm 0.312}$ & $58.124_{\pm 0.187}$ & $97.718_{\pm 0.660}$ & $92.481_{\pm 0.171}$ & $86.359_{\pm 0.162}$ \\
MUSE~\cite{wu2024multimodal} & & \checkmark & \checkmark & $91.359_{\pm 0.057}$ & $65.881_{\pm 0.328}$ & $57.224_{\pm 0.277}$ & $97.351_{\pm 0.052}$ & $91.594_{\pm 0.351}$ & $85.650_{\pm 0.335}$ \\
MoSARe~\cite{moradinasab2025towards} & & \checkmark & \checkmark & $91.565_{\pm 0.061}$ & $65.568_{\pm 0.236}$ & $59.566_{\pm 0.289}$ & $97.681_{\pm 0.032}$ & \uline{92.785}$_{\pm 0.207}$ & $86.069_{\pm 0.236}$ \\
\midrule
\textbf{HP} & \checkmark & \checkmark & \checkmark & $\bm{92.138}_{\pm 0.052}$ & $\bm{68.567}_{\pm 0.381}$ & $\bm{63.367}_{\pm 0.356}$ & $\bm{97.980}_{\pm 0.033}$ & $\bm{93.207}_{\pm 0.103}$ & $\bm{87.203}_{\pm 0.209}$ \\
\bottomrule
\end{tabular}
}
\end{table*}

\vspace{0.2em}
\textit{\textbf{Baselines.}}
In our experiments, we compare our method with 14 recent multimodal methods, each targeting specific types of data incompleteness. These include:
models addressing a \textit{single} type of incompleteness: (1)~MIPM~\cite{zhang2023improving} for irregularly sampled multimodal data; (2)~MEDHMP~\cite{wang2023hierarchical} and VecoCare~\cite{xu2023vecocare} for label sparsity; and (3)~HEART~\cite{huang2024heart}, MuIT-EHR~\cite{chan2024multi}, M3Care~\cite{zhang2022m3care}, DrFuse~\cite{yao2024drfuse}, RedCore~\cite{sun2024redcore}, FlexCare~\cite{xu2024flexcare}, and Diffmv~\cite{zhao2025diffmv} for missing modalities or heterogeneous inputs.
Models tackling \textit{two} types of incompleteness: (4)~PRIME~\cite{li2025prime} for irregular sampling and label sparsity; (5)~UMM~\cite{lee2023learning} for irregular sampling and modality missingness, and (6)~MUSE~\cite{wu2024multimodal} and MoSARe~\cite{moradinasab2025towards} for label and modality missingness.

\vspace{0.2em}
\textit{\textbf{Implementation Details.}}
Our experimental settings are as follows. Hyperparameters in HP are extensively tuned through grid search, and the optimal values are adopted, with parameter sensitivity analyses provided in \ref{app:sensitivity}.

\textit{Data Configuration.} For the time series modality $m^{(1)}$, both MIMIC-III and IV contain 220 time steps. 
\revision{
The time-series modality $m^{(1)}$, textual modality $m^{(2)}$, and imaging modality $m^{(3)}$ are encoded by a two-layer MLP, a fine-tuned Clinical-Longformer~\cite{li2022clinical}, and a frozen DenseNet~\cite{cohen2020limits}, respectively.
}

\textit{Model Settings.} 
\revision{For the three modalities, the initial encoder dimensions $d_{m^{(1)}}^{enc}$, $d_{m^{(2)}}^{enc}$, $d_{m^{(3)}}^{enc}$ are 128, 768, and 1024, respectively. After the first two LRRL layers, modality features dimension are  unified, where $d^u$ is set to 128 for MIMIC-III and 384 for MIMIC-IV.}
The rank $R$ in LRRL is set to 8 across all modalities, and all LRRL and LRRSL modules use 8 attention heads.
\revision{
The local window $\delta$ is set to 2 hours with up to 6 events on MIMIC-III and 48 hours on MIMIC-IV.
}
For LRRSL, we set the sampling intervals $(\Delta t_{m^{(1)}}^1,\Delta t_{m^{(1)}}^3)=(1,4)$ and $(\Delta t_{m^{(2)}}^1,\Delta t_{m^{(2)}}^3)=(4,12)$ hours on MIMIC-III, and $(\Delta t_{m^{(1)}}^1,\Delta t_{m^{(1)}}^3)=(1,4)$ and $(\Delta t_{m^{(3)}}^1,\Delta t_{m^{(3)}}^3)=(12,12)$ hours on MIMIC-IV.
\revision{
Within each modality's sampling window, the sampled token count is determined by $\Delta t_m^l$. Dataset-specific observation windows are described in \ref{appendix:datasets}.
}
In MIMIC-IV, each patient has a single truncated discharge-summary representation ($m^{(2)}$). We exclude this representation from temporal sampling and FGA-based temporal alignment because it does not form an event sequence.

\textit{Loss Weights.} In MIMIC-III, $\lambda_a$ and $\lambda_r$ are set to 0.002 and 10; in MIMIC-IV, they are set to 0.0001 and 5. These scaling factors ensure that different loss components remain on a comparable scale during optimization.

\textit{Optimization.} We adopt the AdamW optimizer~\cite{loshchilov2019decoupled}. All experiments are repeated three times on four NVIDIA H200 GPUs, and we report averaged results along with standard deviations. Further details are provided in \ref{app:implementation}.

\subsection{\textbf{Main Performance}}

Herein, we evaluate the performance of various baselines and our proposed \textbf{HP} on two EHR datasets to answer two core questions:

\begin{itemize} 
    \item \textbf{RQ1:} Can HP enhance IHM prediction performance under multi-level incomplete EHR conditions? 
    \item \textbf{RQ2:} Does HP maintain its superiority as the degree of incompleteness varies? 
\end{itemize}

Notably, all reported results are multiplied by 100. The best results are highlighted in \textbf{bold}, while the second-best are \underline{underlined}.

\subsubsection{\textbf{HP Performance.}}
\label{sec:hp_performance}

To answer \textbf{RQ1}, we report performance under the \textit{Main Experiment} setting (irregular sampling, modality missingness—56\% on MIMIC-III and 85\% on MIMIC-IV, and 50\% label sparsity), as shown in \tabref{tab:main-results}. We observe the following:

HP achieves consistent improvements across all metrics over all baselines.
We attribute this success to the \textit{Clinical Point Cloud Paradigm} and \textit{Low-Rank Relational Attention}, which establish the foundation for interactions among arbitrary clinical events. Building upon this basis, HP achieves fine-grained heterogeneous event fusion, robust modality recovery, and deep self-supervision, enabling it to simultaneously resolve the challenges posed by these three forms of incompleteness, which existing baselines address only partially, as marked in \tabref{tab:main-results}. Specifically, key advantages include:

\textbf{i) Event-level Interaction:} By modeling raw clinical events directly, HP naturally accommodates the structural heterogeneity caused by \textit{irregular sampling} and \textit{missing modalities}. Meanwhile, this paradigm enables fine-grained disease evolution modeling, thereby providing more accurate predictive representations.

\textbf{ii) Robust Modality Recovery:} Unlike single compensation strategies (e.g., M3Care’s similar-case-based recovery or RedCore’s available-modality-based reconstruction), HP integrates these strengths. We recover missing modalities by fusing available \textit{within-case} modalities with \textit{cross-case} priors. 
Furthermore, we employ adaptive entropy-based inference to prioritize high-confidence predictions, mitigating noise from uncertain recovery.

\textbf{iii) Fine-grained Self-supervision:} Compared to baselines relying on coarse-grained (e.g., modality-level) constraints like VecoCare, HP establishes fine-grained, event-level evolution supervision via FGA and FGR. This enables deeper utilization of \textit{unlabeled data} while simultaneously mitigating \textit{temporal irregularity} via alignment and \textit{missing modalities} via reconstruction.


\begin{figure}[!bp]
    \centering
    \includegraphics[width=1.00\columnwidth]{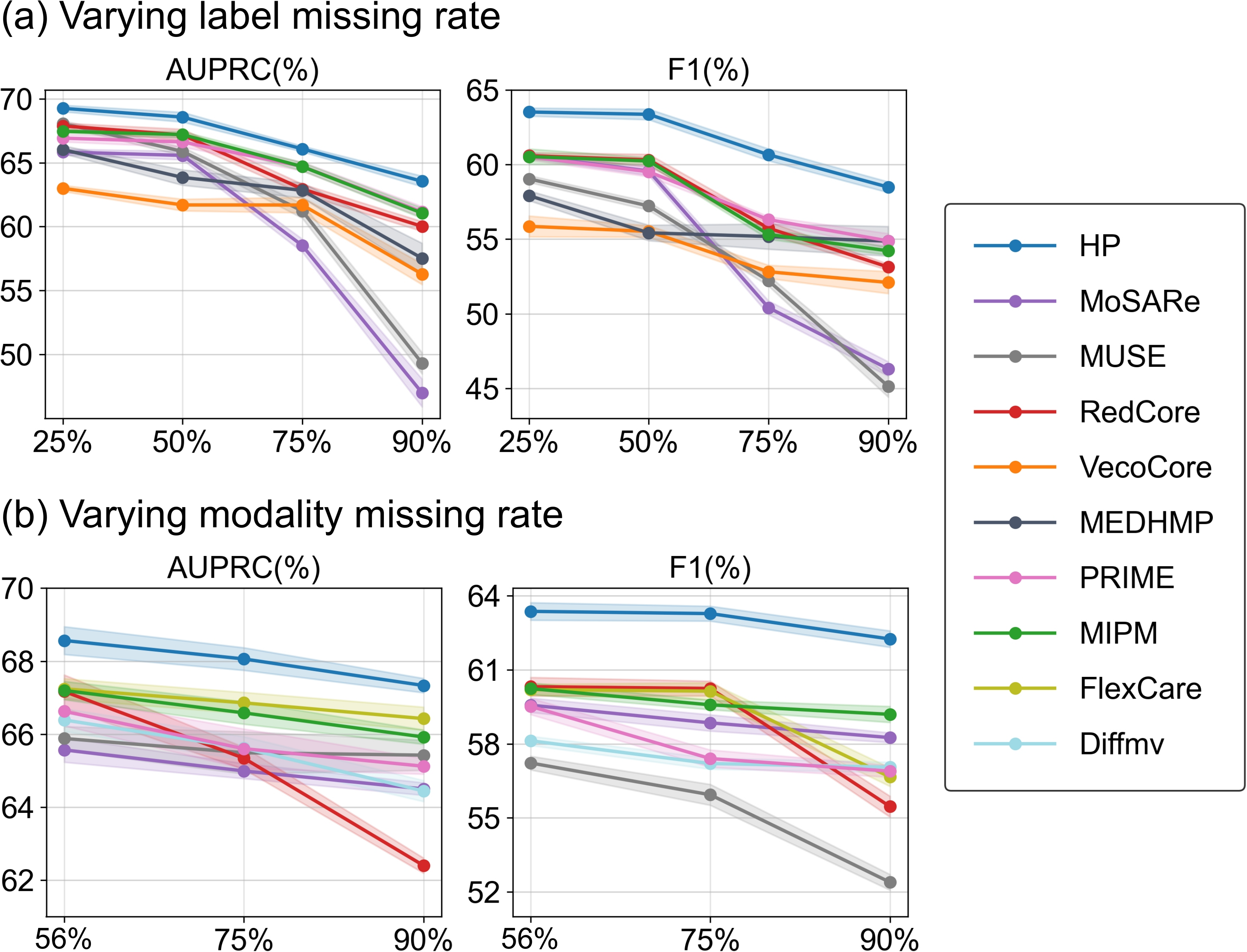}
    \caption{Robustness analysis under varying missing rate.
    } 
    \label{fig:diff_setting}
\end{figure}

\subsubsection{\textbf{Robustness Analysis.}}
\label{sec:robust_analysis}

To answer \textbf{RQ2}, we evaluate robustness of HP by varying label missing rates (25/50/75/90\%) and modality missing rates (56/75/90\%) on MIMIC-III dataset. The comparative results of HP and representative baselines are visualized in \figref{fig:diff_setting}. 
As illustrated, HP (blue line) maintains a significant margin even under extreme conditions (e.g., 90\% missingness).
This demonstrates the high adaptability of the point cloud paradigm and the efficacy of our self-supervised objectives in sparse data regimes.

We further validate HP under decoupled settings: \textit{Only Modality Missing} and \textit{Only Label Missing}. In these experiments, we compare HP against specialized baselines for each setting, as shown in \tabref{tab:only_modality_missing} and \tabref{tab:only_label_missing}. HP remains the top performer, ruling out interference from compounding incomplete factors. 
These results substantiate our analysis in Section \ref{sec:hp_performance}, validating the efficacy of fusing available modalities with cross-case priors for missing modality recovery, and demonstrating the power of fine-grained self-supervision in deeply leveraging sparse labeled data.

\begin{table}[t]
    \centering
    \caption{Performance on the \textit{Only Modality Missing} setting.}
    \label{tab:only_modality_missing}
    \resizebox{\columnwidth}{!}{
        \begin{tabular}{lccccccc}
            \toprule
            \textbf{Metric} & MIPM & RedCore & FlexCare & Diffmv & MUSE & MoSARe & \textbf{HP} \\
            \midrule
            AUROC & 92.085 & 92.168 & 92.113 & 91.821 & 92.178 & \underline{92.270} & \textbf{92.557} \\
            AUPRC & 69.448 & 68.148 & \underline{69.943} & 68.674 & 69.568 & 68.032 & \textbf{70.015} \\
            F1    & \underline{62.840} & 60.632 & 62.410 & 59.633 & 62.352 & 60.765 & \textbf{64.133} \\
            \bottomrule
        \end{tabular}
    }
\end{table}

\begin{table}[t]
    \centering
    \caption{Performance on the \textit{Only Label Missing} setting.}
    \label{tab:only_label_missing}
    \resizebox{\columnwidth}{!}{
        \begin{tabular}{lccccccc}
            \toprule
            \textbf{Metric} & MIPM & PRIME & MEDHMP & VecoCare & MUSE & MoSARe & \textbf{HP} \\
            \midrule
            AUROC & 82.821 & 82.971 & 85.106 & 82.167 & 80.942 & \underline{85.640} & \textbf{85.686} \\
            AUPRC & 42.707 & 42.698 & 42.234 & 42.043 & 38.133 & \underline{45.065} & \textbf{51.414} \\
            F1    & 40.237 & 41.282 & 40.538 & \underline{43.088} & 38.565 & 39.021 & \textbf{51.301} \\
            \bottomrule
        \end{tabular}
    }
\end{table}

\begin{figure}[!tbp]
    \centering
    \includegraphics[width=1.00\columnwidth]{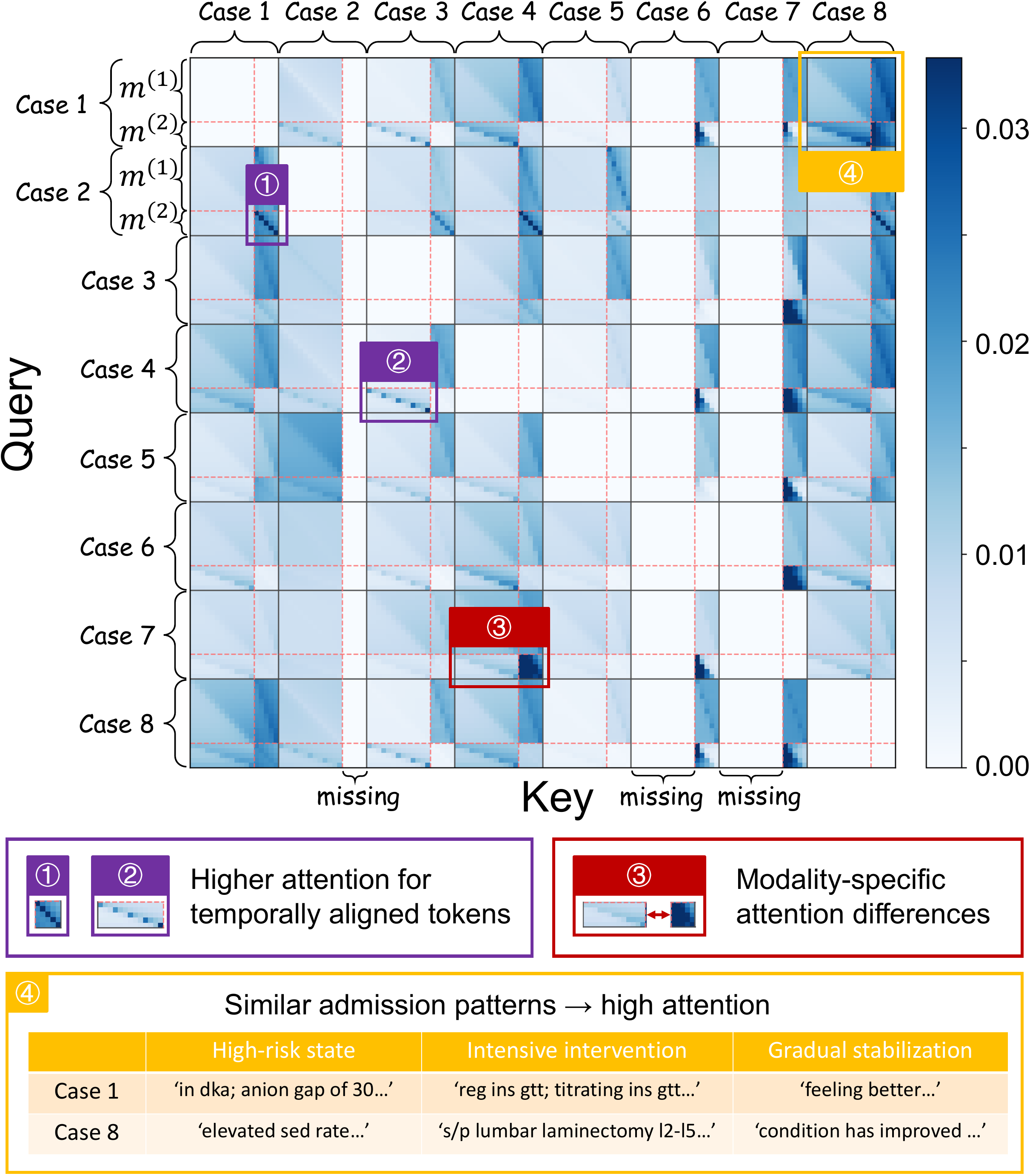}
    \caption{Case Study. 
    } 
    \label{fig:case_study}
\end{figure}

\subsection{\textbf{Case Study}}
\label{sec:case_study}

The key component of our clinical point cloud paradigm is LRRL, which enables interaction modeling between arbitrary point pairs via relative relation learning. 
To examine its effectiveness in jointly coupling content, time, modality, and case dimensions, we visualize the attention logits of the \textit{Cross-Case LRRL} in Figure~\ref{fig:case_study}. 
We analyze dependencies across 8 cases, each containing two modalities ($m^{(1)}$: 13 steps; $m^{(2)}$: 5 steps). 
The heatmap reveals three key patterns:

i) \textit{Time Dimension}: Regions \ding{192} and \ding{193} show higher attention for \textit{temporally aligned tokens} regardless of modality.
This indicates that LRRL is sensitive to temporal factors and tends to attend to disease states at synchronized admission stages in other cases.

ii) \textit{Modality Dimension}: As seen in \ding{194}, cross-case interactions prioritize same-modality pairs (e.g., $m^{(2)}$-$m^{(2)}$), confirming that the modality dimension effectively distinguishes and preserves modality-specific semantics.

iii) \textit{Case Dimension}: Region \ding{195} highlights strong dependencies between Case 1 and Case 8. This corresponds to their semantically similar trajectories (both exhibiting High-risk $\rightarrow$ Intervention $\rightarrow$ Stabilization), demonstrating that LRRL effectively quantifies high-order patient case similarity to leverage historical priors.

\begin{figure}[!bp]
    \centering
    \includegraphics[width=1.00\columnwidth]{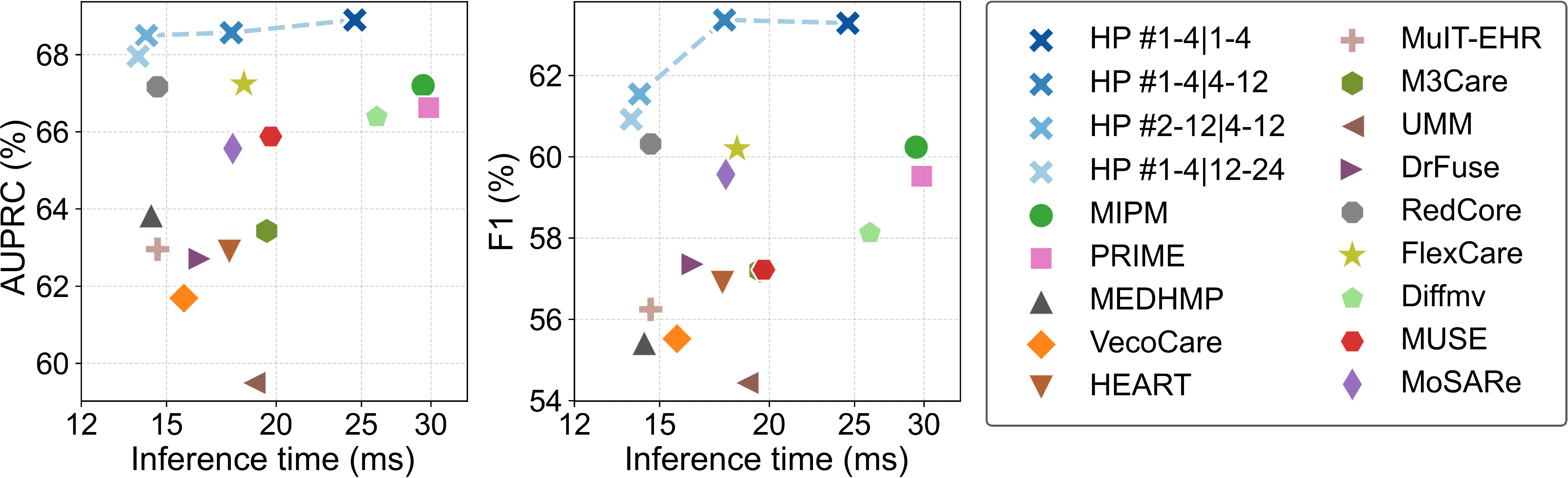}
    \caption{Performance vs. Inference Time. 
    }
    \label{fig:cost}
\end{figure}

\subsection{\textbf{Cost Analysis}}
To evaluate computational cost and validate the efficiency-granularity balance of our \textit{Low-Rank Relational Sampled Layer} (LRRSL), Figure~\ref{fig:cost} visualizes inference time versus performance (AUPRC/F1) for both HP and baselines. Here, HP is evaluated across varying sampling configurations, denoted as ``HP \#$\Delta t^1_{m^{(1)}}$-$\Delta t^3_{m^{(1)}}$ \textbar\ $\Delta t^1_{m^{(2)}}$-$\Delta t^3_{m^{(2)}}$''.
As shown in Figure~\ref{fig:cost}, three observations can be drawn:
1) Increasing sampling intervals significantly reduces inference latency, confirming that our design effectively prunes computations.
2) Overly coarse sampling leads to performance degradation, highlighting the importance of fine-grained temporal modeling for capturing disease evolution patterns.
3) The configuration ``HP \#1-4 \textbar\ 4-12'' achieves an optimal trade-off, maintaining top-tier performance at competitive computational costs. This demonstrates that our \textit{Hierarchical Interaction and Sampling} strategy achieves an effective balance.

\subsection{\textbf{Ablation Study}}
Herein, to validate the \textit{low-rank relational attention} and \textit{self-supervised strategy}, ablation studies are conducted on MIMIC-III. Results are shown in \tabref{tab:ablation}, with supplementary analyses in \ref{app:ablation_details}.

\textbf{i) Low-rank Relational Mechanism.}
We systematically ablate each coordinate dimension (e.g., ``w/o time'') to evaluate their individual contributions. Additionally, to validate our low-rank coupling strategy, we replace it with element-wise summation (``SUM'') or concatenation (``Concat''). Performance degradation across all variants confirms two key insights: 1) all four dimensions are indispensable for characterizing clinical event correlations; and 2) the proposed low-rank mechanism is superior in coupling multi-dimensional features and measuring high-order dependencies between arbitrary point pairs.

\textbf{ii) Self-supervision Strategy.}
We assess our self-supervised objectives by removing Fine-grained Alignment (``w/o FGA''), Reconstruction (``w/o FGR''), or both. The resulting performance drops justify the synergy between contrastive alignment and reconstruction constraints. Furthermore, degrading the supervision to coarse modality-level representations (``w/o fine-grained'') causes significant decline, demonstrating that fine-grained, event-level supervision is crucial for capturing patient condition dynamics and maximizing the utility of sparse labels.


\begin{table}[hb]
\centering
\caption{Ablation study.}
\label{tab:ablation}
\setlength{\tabcolsep}{11pt} 
\resizebox{\linewidth}{!}{
\begin{tabular}{lccc}
\toprule
\textbf{Variant} & \textbf{AUROC (\%)} & \textbf{AUPRC (\%)} & \textbf{F1 (\%)} \\
\midrule
SUM &
$91.780_{\pm0.048}$ &
$67.809_{\pm0.390}$ &
$62.008_{\pm0.337}$ \\
Concat &
$91.775_{\pm0.059}$ &
$68.091_{\pm0.375}$ &
$62.580_{\pm0.369}$ \\
w/o content &
$91.385_{\pm0.023}$ &
$66.899_{\pm0.290}$ &
$61.039_{\pm0.278}$ \\
w/o time &
$91.459_{\pm0.039}$ &
$66.398_{\pm0.273}$ &
$59.936_{\pm0.401}$ \\
w/o modality &
$91.630_{\pm0.028}$ &
$67.573_{\pm0.355}$ &
$61.237_{\pm0.425}$ \\
w/o case &
$91.593_{\pm0.032}$ &
$67.747_{\pm0.219}$ &
$61.893_{\pm0.365}$ \\
\midrule
w/o FGA &
$91.823_{\pm0.055}$ &
$67.784_{\pm0.276}$ &
$61.593_{\pm0.317}$ \\
w/o FGR &
$91.926_{\pm0.031}$ &
$67.546_{\pm0.258}$ &
$61.427_{\pm0.290}$ \\
w/o FGA+FGR &
$91.653_{\pm0.037}$ &
$66.310_{\pm0.307}$ &
$61.001_{\pm0.388}$ \\
w/o fine-grained &
$91.932_{\pm0.028}$ &
$68.243_{\pm0.357}$ &
$62.936_{\pm0.351}$ \\
\midrule
\textbf{HP (Full)} &
$\bm{92.138}_{\pm0.052}$ &
$\bm{68.567}_{\pm0.381}$ &
$\bm{63.367}_{\pm0.356}$ \\
\bottomrule
\end{tabular}
}
\end{table}

\section{Related work}
\label{section-5}

Multimodal learning integrates heterogeneous signals through cross-modal attention and alignment \cite{tsai2019multimodal, zhang2022contrastive}; healthcare models extend these mechanisms to adaptive fusion and clinical prediction \cite{wang2025moe, king2023multimodal}.
However, real-world EHRs inherently suffer from multi-level incompleteness \cite{johnson2016mimic, wu2024multimodal}, including \textit{irregular sampling}, \textit{missing modalities}, and \textit{label scarcity}, which challenges models assuming data completeness. Recent research addresses these issues as follows:

\textit{\textbf{Irregular Sampling}} disrupts the temporal alignment of disease progression representations. Representative uni-modal approaches use recurrent models with missingness information~\cite{che2018recurrent}, continuous-time attention~\cite{chen2024contiformer}, or multi-scale temporal warping~\cite{zhang2023warpformer}. Multimodal settings additionally require fusion across asynchronous timelines. Existing solutions employ \textit{cross-modal alignment}~\cite{wang2025ctpd, li2025prime, zhang2023improving} or unify observations into \textit{time-aware tokens} to bypass explicit alignment~\cite{lee2023learning}.

\textit{\textbf{Missing Modalities}} lead to severe modality imbalance during fusion. Existing strategies generally fall into three categories: 1)~\textit{Structural Adaptation}, which explicitly ignores missing inputs~\cite{yao2024drfuse, lee2023learning, xu2024flexcare}; 2)~\textit{Self-Reconstruction}, which imputes missing views from available ones~\cite{sun2024redcore, zhao2025diffmv}; and 3)~\textit{Similar-Case Retrieval}, which leverages priors from similar cases for recovery~\cite{zhang2022m3care, zhi2025borrowing, lang2025redeeming, li2026learning}.

\textit{\textbf{Label Scarcity}} hinders robust learning due to limited supervision. To address this, Self-Supervised Learning (SSL) is widely adopted to exploit intrinsic data constraints.
While early works treated alignment and reconstruction independently~\cite{zhang2022contrastive, zhang2023multi}, recent advances have begun to integrate both techniques \cite{xu2023vecocare, wang2023hierarchical, king2023multimodal}. PRIME~\cite{li2025prime} further refines this by advancing from coarse modality-level to fine-grained evolution-level alignment.

Crucially, most existing models address these issues in isolation or at most in pairs. 
When all three levels of incompleteness coexist, models are forced into rigid alignment, sample exclusion, or decoupled unimodal encoding that impedes fine-grained fusion, causing clinical information loss.
In response, we propose HealthPoint (HP), which simultaneously resolves this tripartite challenge within a cohesive \textit{Clinical Point Cloud Paradigm}. 
Note that we focus on raw heterogeneous observations, distinct from research targeting structured clinical entities or predefined codes~\cite{choi2018mime, huang2024heart, chan2024multi}.

\section{Conclusion}
\label{sec:conclusion}

We propose HealthPoint (HP), a clinical point cloud paradigm for event-level multimodal fusion in mortality prediction under multi-level incomplete EHRs.
Clinical events are represented within a quasi-4D coordinate system defined by content, time, modality, and patient case.
Low-rank relational attention couples dependencies across these dimensions, while hierarchical interaction and sampling balance representation granularity with computational efficiency.
Fine-grained alignment and reconstruction objectives support missing-modality recovery and representation learning from unlabeled observations.
Experiments on MIMIC-III and MIMIC-IV show improved predictive performance over the evaluated baselines, supported by missingness analyses, ablations, and efficiency comparisons.
These findings support event-level relational fusion as a promising approach to robust healthcare representation learning.
The evidence comes from retrospective EHR benchmarks and simulated missingness settings. In particular, MIMIC-IV uses discharge-aligned imaging; its results do not establish early-warning performance at a fixed landmark after admission. Prospective evaluation at clinically actionable prediction times and validation across healthcare institutions are needed to assess clinical utility and generalizability.

\appendix

\section{Theoretical Justification of Low-Rank Coupling}
\label{app:proof}

In this section, we show that the proposed \textit{Low-Rank Coupling} (Eq.~\ref{eq:low_rank}) is a CP-based low-rank approximation of full high-order interactions among heterogeneous clinical dimensions~\cite{kolda2009tensor}.

\noindent \textbf{Full interaction.}
For a clinical point pair $(i,j)$ with relational features $\mathcal{R}_{ij}=\{\bm{r}_{ij}^h,\bm{r}_{ij}^t,\bm{r}_{ij}^m,\bm{r}_{ij}^c\}$ over $D=4$ dimensions, the ideal interaction is
\begin{equation}
e_{ij}^{\text{ideal}}
=
\langle
\mathcal{W},
\bm{r}_{ij}^h \otimes \bm{r}_{ij}^t \otimes \bm{r}_{ij}^m \otimes \bm{r}_{ij}^c
\rangle
+\text{bias},
\label{eq:ideal_tensor}
\end{equation}
where $\mathcal{W}\in\mathbb{R}^{d\times d\times d\times d}$ is a full weight tensor, requiring $O(d^4)$ parameters and computation.

\noindent \textbf{Low-rank approximation.}
Assuming $\mathcal{W}$ is low-rank, CP decomposition gives
\begin{equation}
\mathcal{W}
\approx
\sum_{\gamma=1}^{R}
\mathbf{Q}_{h}^{(\gamma)} \otimes
\mathbf{Q}_{t}^{(\gamma)} \otimes
\mathbf{Q}_{m}^{(\gamma)} \otimes
\mathbf{Q}_{c}^{(\gamma)},
\label{eq:cp_decomp}
\end{equation}
where $\mathbf{Q}_{*}^{(\gamma)}\in\mathbb{R}^{d}$.

Substituting Eq.~\ref{eq:cp_decomp} into Eq.~\ref{eq:ideal_tensor} yields
\begin{align}
e_{ij}^{\text{coupled}}
&=
\sum_{\gamma=1}^{R}
\left\langle
\bigotimes_{* \in \mathcal{D}} \mathbf{Q}_{*}^{(\gamma)},
\bigotimes_{* \in \mathcal{D}} \bm{r}_{ij}^{*}
\right\rangle \nonumber\\
&=
\sum_{\gamma=1}^{R}
\prod_{* \in \mathcal{D}}
\langle \mathbf{Q}_{*}^{(\gamma)}, \bm{r}_{ij}^{*} \rangle,
\end{align}
which is exactly the coupled term in Eq.~\ref{eq:low_rank}.

\noindent \textbf{Conclusion.}
Our low-rank coupling is therefore a CP approximation of the full high-order interaction tensor: the coupled term models $D$-th order multiplicative dependencies, while the unary term captures first-order linear effects. This reduces the complexity from $O(d^D)$ to $O(RdD)$.

\section{Experimental settings}
\subsection{Dataset Description and Preprocessing}
\label{appendix:datasets}

We use two large-scale multimodal EHR datasets: MIMIC-III and MIMIC-IV. MIMIC-III contains \revision{irregularly sampled multivariate time series} ($m^{(1)}$) and clinical note sequences ($m^{(2)}$). MIMIC-IV includes $m^{(1)}$, truncated discharge summaries ($m^{(2)}$), and irregularly sampled chest X-ray sequences ($m^{(3)}$). Below we summarize preprocessing, dataset statistics, and incomplete-data simulation.

\subsubsection{\textbf{Data Preprocessing}}

\textbf{MIMIC-III.}
We construct the in-hospital mortality (IHM) dataset using official scripts~\cite{johnson2016mimic}. The \revision{irregularly sampled  time series} ($m^{(1)}$) are extracted with the benchmark pipeline~\cite{harutyunyan2019multitask}, and irregular clinical note sequences ($m^{(2)}$) are built following~\cite{khadanga2019using}. The two modalities are merged as in~\cite{zhang2023improving}, retaining partially observed samples. Only the first 48 hours after admission are used.

\textbf{MIMIC-IV.}
This dataset includes time series ($m^{(1)}$), discharge summaries ($m^{(2)}$), and chest X-ray sequences ($m^{(3)}$). Data are collected from MIMIC-IV~\cite{johnson2023mimic}, MIMIC-IV-Note~\cite{PhysioNet-mimic-iv-note-2.2}, and MIMIC-CXR~\cite{PhysioNet-mimic-cxr-jpg-2.1.0}. Time series are extracted using the same pipeline~\cite{harutyunyan2019multitask}. To limit outcome-bearing information in discharge summaries, we retain only \textit{Chief Complaint}, \textit{Medication on Admission}, and \textit{Past Medical History}~\cite{lee2023learning}. The prediction landmark is the discharge time, denoted by $t_{\mathrm{discharge}}$. Chest X-rays acquired in $[t_{\mathrm{discharge}}-48\,\mathrm{h},\,t_{\mathrm{discharge}}]$ are used as $m^{(3)}$; thus, the imaging window covers the 48 hours preceding both discharge and prediction. This discharge-aligned protocol evaluates retrospective mortality discrimination and does not establish performance at a fixed early prediction landmark after admission.

\revision{All modalities preserve their original observation timestamps and are therefore irregularly sampled.}
All time-series features are normalized, and each text segment is truncated to 512 tokens.

\subsubsection{\textbf{Data Statistics}}

\begin{table*}[t]
\centering
\caption{Train/val/test split statistics for MIMIC-III and MIMIC-IV under various incompleteness settings. Percentages in parentheses are rounded to the nearest whole percent.}
\label{tab:data-splits}
\resizebox{\textwidth}{!}{%
\begin{tabular}{l|ccc|ccc|ccc}
\toprule
\multirow{2}{*}{\textbf{Setting}} & \multicolumn{3}{c|}{\textbf{Train}} & \multicolumn{3}{c|}{\textbf{Val}} & \multicolumn{3}{c}{\textbf{Test}} \\
 & Total & Label Missing & Mod Missing & Total & Label Missing & Mod Missing & Total & Label Missing & Mod Missing \\
\midrule
\multicolumn{10}{l}{\textbf{MIMIC-III}} \\
Raw & 25172 & 0 (0\%) & 14214 (56\%) & 6293 & 0 & 3596 & 5556 & 0 & 3068 \\
Main Experiment & 25172 & 12586 (50\%) & 14214 (56\%) & 6293 & 0 & 3596 & 5556 & 0 & 3068 \\
25\% Label Missing & 25172 & \textbf{6293 (25\%)} & 14214 (56\%) & 6293 & 0 & 3596 & 5556 & 0 & 3068 \\
50\% Label Missing & 25172 & \textbf{12586 (50\%)} & 14214 (56\%) & 6293 & 0 & 3596 & 5556 & 0 & 3068 \\
75\% Label Missing & 25172 & \textbf{18879 (75\%)} & 14214 (56\%) & 6293 & 0 & 3596 & 5556 & 0 & 3068 \\
90\% Label Missing & 25172 & \textbf{22654 (90\%)} & 14214 (56\%) & 6293 & 0 & 3596 & 5556 & 0 & 3068 \\
56\% Modality Missing & 25172 & 12586 (50\%) & \textbf{14214 (56\%)} & 6293 & 0 & 3596 & 5556 & 0 & 3068 \\
75\% Modality Missing & 25172 & 12586 (50\%) & \textbf{18879 (75\%)} & 6293 & 0 & 3596 & 5556 & 0 & 3068 \\
90\% Modality Missing & 25172 & 12586 (50\%) & \textbf{22655 (90\%)} & 6293 & 0 & 3596 & 5556 & 0 & 3068 \\
\revision{Only Modality Missing} & 25172 & \textbf{0 (0\%)} & 14214 (56\%) & 6293 & 0 & 3596 & 5556 & 0 & 3068 \\
\revision{Only Label Missing} & 10958 & 9862 (90\%) & \textbf{0 (0\%)} & 2697 & 0 & 0 & 2488 & 0 & 0 \\
\midrule
\multicolumn{10}{l}{\textbf{MIMIC-IV}} \\
Raw & 22033 & 0 (0\%) & 18795 (85\%) & 5445 & 0 & 4658 & 3408 & 0 & 2745 \\
Main Experiment & 22033 & 11016 (50\%) & 18795 (85\%) & 5445 & 0 & 4658 & 3408 & 0 & 2745 \\
\bottomrule
\end{tabular}%
}
\end{table*}

\begin{table*}[t]
\centering
\caption{Modality-specific missingness statistics under the main experiment setting.}
\label{tab:modality-missing}
\resizebox{0.9\textwidth}{!}{%
\begin{tabular}{lcccccccccccc}
\toprule
\multirow{2}{*}{\textbf{Dataset}} & \multicolumn{4}{c}{\textbf{Train}} & \multicolumn{4}{c}{\textbf{Val}} & \multicolumn{4}{c}{\textbf{Test}} \\
\cmidrule(lr){2-5} \cmidrule(lr){6-9} \cmidrule(lr){10-13}
 & \textbf{Total} & \textbf{$m^{(1)}$ miss} & \textbf{$m^{(2)}$ miss} & \textbf{$m^{(3)}$ miss}
 & \textbf{Total} & \textbf{$m^{(1)}$ miss} & \textbf{$m^{(2)}$ miss} & \textbf{$m^{(3)}$ miss}
 & \textbf{Total} & \textbf{$m^{(1)}$ miss} & \textbf{$m^{(2)}$ miss} & \textbf{$m^{(3)}$ miss} \\
\midrule
MIMIC-III & 25172 & 3394 & 10820 & -- & 6293 & 2742 & 854 & -- & 5556 & 2320 & 748 & -- \\
MIMIC-IV  & 22033 & 0 & 6070 & 18752 & 5445 & 0 & 1435 & 4650 & 3408 & 0 & 174 & 2741 \\
\bottomrule
\end{tabular}
}
\end{table*}

The multivariate time series ($m^{(1)}$) modality contains 17 clinical variables, including \textit{capillary refill rate, blood pressures, oxygen metrics, glucose, GCS scores, heart rate, temperature}, among others. 
MIMIC-III clinical notes ($m^{(2)}$) are collected from nursing and physician reports, providing rich contextual data on patient status. In MIMIC-IV, we restrict $m^{(2)}$ to a few pre-admission fields to minimize target leakage. Chest X-rays ($m^{(3)}$) are irregularly sampled and consist of both frontal and lateral views.

To simulate label sparsity, we randomly remove 50\% of labels in the training set as our \textit{main experimental} condition. 
To assess robustness under various types and degrees of incompleteness, we additionally construct the following settings based on either the \textit{raw dataset} or by further dropping labels/modalities from the \textit{main experimental} dataset:
\begin{enumerate}
    \item \textbf{Varying label missing ratios:} 25\%, 50\%, 75\%, and 90\%.
    \item \textbf{Varying modality missing ratios:} 56\%, 75\%, and 90\%.
    \item \textbf{Only modality missing:} Labels fully observed, modality missing only.
    \item \textbf{Only label missing:} All modalities present, labels sparsely available.
\end{enumerate}

The data splits and missingness statistics for each setting across MIMIC-III and MIMIC-IV are summarized in Table~\ref{tab:data-splits}.
We further report the modality-specific missingness statistics under our main experimental setting in Table~\ref{tab:modality-missing}.

\subsection{Baseline Models}
\label{appendix:baselines}

We compare our model against 14 recent multimodal models, each designed to handle different types of incompleteness in EHRs. 
To ensure a fair comparison, all baselines are evaluated under a consistent data configuration.
And, we prioritize preserving the original architectural designs of all baselines.
However, when a baseline lacks native support for specific modalities (e.g., imaging), we employ a unified implementation to minimize performance variance caused by encoder differences:
\begin{itemize}
    \item \textbf{Time series}: Missing values are filled using backward imputation, and irregular sampling is addressed with UTDE~\cite{zhang2023improving}.
    \item \textbf{Text}: Each clinical note is encoded using ClinicalLongformer, then aggregated via an RNN/Transformer.
    \item \textbf{Imaging}: Imaging features are extracted with DenseNet and sequentially modeled using an RNN/Transformer.
\end{itemize}


\begin{table}[tb]
\centering
\caption{Batch size and loss weight settings.}
\label{tab:training-config}
\resizebox{0.85\columnwidth}{!}{%
\begin{tabular}{l|cc|cc}
\toprule
\textbf{Setting} & \textbf{Train Batch} & \textbf{Infer Batch} & $\boldsymbol{\lambda_a}$ & $\boldsymbol{\lambda_r}$ \\
\midrule
\multicolumn{5}{l}{\textbf{MIMIC-III}} \\
Main Experiment & 16 & 32 & 0.002 & 10 \\
25\% Label Missing & 16 & 32 & 0.02 & 10 \\
50\% Label Missing & 16 & 32 & 0.002 & 10 \\
75\% Label Missing & 16 & 32 & 0.002 & 10 \\
90\% Label Missing & 16 & 32 & 0.002 & 10 \\
56\% Modality Missing & 16 & 32 & 0.002 & 10 \\
75\% Modality Missing & 16 & 32 & 0.02 & 10 \\
90\% Modality Missing & 16 & 32 & 0.02 & 10 \\
Only Modality Missing & 16 & 32 & 0.02 & 10 \\
Only Label Missing & 16 & 32 & 0.02 & 10 \\
\midrule
\multicolumn{5}{l}{\textbf{MIMIC-IV}} \\
Main Experiment & 32 & 128 & 0.0001 & 5 \\
\bottomrule
\end{tabular}%
}
\end{table}


\subsection{Training Configuration} 
\label{app:implementation}

We provide additional implementation details omitted from the main text. For the temporal window $\delta$ in the first LRRL, we set $\delta=2$ hours with up to 6 clinical events for MIMIC-III, and use a 48-hour window for MIMIC-IV. All LRRL and LRRSL modules use 8 attention heads. The learning rates are fixed at 2e-5 for BERT-based modules and 8e-4 for all other components. The training/inference batch sizes and loss weights ($\lambda_a$, $\lambda_r$) under different settings are summarized in Table~\ref{tab:training-config}. HP is trained for 30 epochs on MIMIC-III and 10 epochs on MIMIC-IV.
We use a larger $\lambda_a$ under more severe incompleteness, while a smaller $\lambda_a$ is adopted on MIMIC-IV due to its stronger cross-modal asynchrony.

\section{Experimental Results Analysis}

\subsection{Performance Comparison with Baselines}

We further compare HP with different baseline categories to clarify the source of its performance gains.


\textit{i) Compared with methods designed mainly for temporal irregularity}, HP further models modality missingness and label sparsity, leading to more robust representations under realistic incomplete EHR settings.

\textit{ii) Compared with methods focused on label sparsity}, HP uses finer-grained event-level self-supervision and is simultaneously compatible with temporal irregularity and modality imbalance, allowing more effective use of unlabeled data.

\textit{iii) Comparison with methods for missing modalities}, HP combines structural adaptability to missingness with modality recovery from both within-case multimodal cues and cross-case priors. The entropy-based inference strategy further improves robustness by reducing the impact of uncertain recovered representations.

\textit{iv) Compared with methods that address only part of the incompleteness}, HP provides a unified solution to the coupled challenges of irregularity, modality missingness, and label sparsity, resulting in more stable and effective representations.

\begin{table}[t]
    \centering
    \caption{\revision{Paired bootstrap comparison between HP and RedCore.}}
    \label{tab:statistical_significance}
    \small
    \setlength{\tabcolsep}{4pt}
    \resizebox{\columnwidth}{!}{%
\begin{tabular}{l|ccc}
\toprule
        Dataset 
        & AUROC $\Delta$ [95\% CI]
        & AUPRC $\Delta$ [95\% CI]
        & F1 $\Delta$ [95\% CI] \\
        \midrule
        MIMIC-III
        & $+0.52\;[-0.17,\;1.25]$
        & $+1.35\;[0.16,\;3.26]^{*}$
        & $+3.02\;[1.28,\;6.08]^{*}$ \\
        
        MIMIC-IV
        & $+0.17\;[-0.19,\;0.59]$
        & $+0.56\;[-0.36,\;1.53]$
        & $+0.66\;[0.11,\;2.52]^{*}$ \\
        \bottomrule
        \multicolumn{4}{l}{
            \footnotesize $^{*}$ The 95\% confidence interval excludes zero.
        }
\end{tabular}%
}
\end{table}

\textit{\revision{\textbf{Statistical significance analysis}}}. \revision{We performed 10,000 patient-level paired stratified bootstrap resamples, each drawn with replacement to match the original test-set size. Compared with RedCore, a strong competing baseline, HP achieved positive gains across all six metrics, with three reaching statistical significance; absolute gains and 95\% bootstrap confidence intervals are reported in Table \ref{tab:statistical_significance}.}

\begin{table}[t]
    \centering
    \caption{Performance comparison under varying label missing rates on MIMIC-III dataset.}
    \label{tab:label_robustness}
    \resizebox{0.9\columnwidth}{!}{ 
    \begin{tabular}{lcccc}
        \toprule
        Method & Missing Rate & AUROC & AUPRC & F1 \\
        \midrule
        \multirow{4}{*}{MIPM} 
        & 25\% & $91.796_{\pm 0.023}$ & $67.457_{\pm 0.119}$ & $60.534_{\pm 0.517}$ \\
        & 50\% & $91.621_{\pm 0.041}$ & $67.197_{\pm 0.252}$ & $60.239_{\pm 0.236}$ \\
        & 75\% & $90.718_{\pm 0.083}$ & $64.689_{\pm 0.326}$ & $55.319_{\pm 0.319}$ \\
        & 90\% & $89.350_{\pm 0.192}$ & $61.056_{\pm 0.403}$ & $54.219_{\pm 0.282}$ \\
        \midrule
        \multirow{4}{*}{PRIME} 
        & 25\% & $91.767_{\pm 0.029}$ & $66.920_{\pm 0.287}$ & $60.531_{\pm 0.218}$ \\
        & 50\% & $91.537_{\pm 0.066}$ & $66.625_{\pm 0.394}$ & $59.518_{\pm 0.329}$ \\
        & 75\% & $90.725_{\pm 0.071}$ & $64.702_{\pm 0.347}$ & $56.311_{\pm 0.208}$ \\
        & 90\% & $89.435_{\pm 0.095}$ & $61.153_{\pm 0.428}$ & $54.882_{\pm 0.496}$ \\
        \midrule
        \multirow{4}{*}{MEDHMP} 
        & 25\% & $91.389_{\pm 0.035}$ & $66.023_{\pm 0.310}$ & $57.918_{\pm 0.328}$ \\
        & 50\% & $90.091_{\pm 0.081}$ & $63.842_{\pm 0.603}$ & $55.423_{\pm 0.522}$ \\
        & 75\% & $89.872_{\pm 0.076}$ & $62.836_{\pm 0.581}$ & $55.178_{\pm 0.847}$ \\
        & 90\% & $88.877_{\pm 0.129}$ & $57.518_{\pm 1.183}$ & $54.866_{\pm 0.992}$ \\
        \midrule
        \multirow{4}{*}{VecoCare} 
        & 25\% & $90.362_{\pm 0.048}$ & $62.992_{\pm 0.237}$ & $55.861_{\pm 0.699}$ \\
        & 50\% & $90.234_{\pm 0.063}$ & $61.692_{\pm 0.457}$ & $55.522_{\pm 0.383}$ \\
        & 75\% & $89.251_{\pm 0.086}$ & $61.686_{\pm 0.605}$ & $52.816_{\pm 0.442}$ \\
        & 90\% & $87.481_{\pm 0.157}$ & $56.283_{\pm 0.817}$ & $52.099_{\pm 0.739}$ \\
        \midrule
        \multirow{4}{*}{RedCore} 
        & 25\% & $92.113_{\pm 0.083}$ & $67.876_{\pm 0.226}$ & $60.593_{\pm 0.212}$ \\
        & 50\% & $91.710_{\pm 0.069}$ & $67.169_{\pm 0.455}$ & $60.316_{\pm 0.377}$ \\
        & 75\% & $90.934_{\pm 0.106}$ & $62.936_{\pm 0.387}$ & $55.738_{\pm 0.425}$ \\
        & 90\% & $89.800_{\pm 0.055}$ & $60.016_{\pm 0.382}$ & $53.132_{\pm 0.203}$ \\
        \midrule
        \multirow{4}{*}{MUSE} 
        & 25\% & $91.691_{\pm 0.036}$ & $68.063_{\pm 0.226}$ & $59.040_{\pm 0.230}$ \\
        & 50\% & $91.359_{\pm 0.057}$ & $65.881_{\pm 0.328}$ & $57.224_{\pm 0.277}$ \\
        & 75\% & $90.135_{\pm 0.112}$ & $61.217_{\pm 0.562}$ & $52.217_{\pm 0.351}$ \\
        & 90\% & $84.620_{\pm 0.185}$ & $49.302_{\pm 0.828}$ & $45.139_{\pm 0.718}$ \\
        \midrule
        \multirow{4}{*}{MoSARe} 
        & 25\% & $91.572_{\pm 0.058}$ & $65.835_{\pm 0.228}$ & $60.606_{\pm 0.182}$ \\
        & 50\% & $91.565_{\pm 0.081}$ & $65.568_{\pm 0.336}$ & $59.566_{\pm 0.289}$ \\
        & 75\% & $88.848_{\pm 0.172}$ & $60.515_{\pm 0.503}$ & $50.409_{\pm 0.457}$ \\
        & 90\% & $85.183_{\pm 0.132}$ & $46.982_{\pm 1.086}$ & $46.305_{\pm 0.482}$ \\
        \midrule
        \multirow{4}{*}{\textbf{HP}} 
        & 25\% & $\mathbf{92.146_{\pm 0.039}}$ & $\mathbf{69.251_{\pm 0.258}}$ & $\mathbf{63.525_{\pm 0.271}}$ \\
        & 50\% & $\mathbf{92.138_{\pm 0.052}}$ & $\mathbf{68.567_{\pm 0.381}}$ & $\mathbf{63.367_{\pm 0.356}}$ \\
        & 75\% & $\mathbf{91.223_{\pm 0.103}}$ & $\mathbf{66.078_{\pm 0.226}}$ & $\mathbf{60.659_{\pm 0.398}}$ \\
        & 90\% & $\mathbf{90.176_{\pm 0.167}}$ & $\mathbf{63.543_{\pm 0.414}}$ & $\mathbf{58.489_{\pm 0.358}}$ \\
        \bottomrule
    \end{tabular}
    }
\end{table}

\begin{table}[t]
    \centering
    \caption{Performance comparison under varying modality missing rates on MIMIC-III dataset.}
    \label{tab:modality_robustness}
    \resizebox{0.9\linewidth}{!}{
    \begin{tabular}{lcccc}
        \toprule
        Method & Missing Rate & AUROC & AUPRC & F1 \\
        \midrule
        \multirow{3}{*}{MIPM} 
        & 56\% & $91.621_{\pm 0.041}$ & $67.197_{\pm 0.252}$ & $60.239_{\pm 0.236}$ \\
        & 75\% & $91.581_{\pm 0.021}$ & $66.583_{\pm 0.308}$ & $59.579_{\pm 0.193}$ \\
        & 90\% & $91.572_{\pm 0.031}$ & $65.922_{\pm 0.195}$ & $59.194_{\pm 0.324}$ \\
        \midrule
        \multirow{3}{*}{PRIME} 
        & 56\% & $91.537_{\pm 0.066}$ & $66.625_{\pm 0.394}$ & $59.518_{\pm 0.329}$ \\
        & 75\% & $91.292_{\pm 0.027}$ & $65.602_{\pm 0.541}$ & $57.403_{\pm 0.350}$ \\
        & 90\% & $91.248_{\pm 0.031}$ & $65.121_{\pm 0.215}$ & $56.893_{\pm 0.239}$ \\
        \midrule
        \multirow{3}{*}{RedCore} 
        & 56\% & $91.710_{\pm 0.069}$ & $67.169_{\pm 0.455}$ & $60.316_{\pm 0.377}$ \\
        & 75\% & $91.592_{\pm 0.052}$ & $65.341_{\pm 0.302}$ & $60.244_{\pm 0.299}$ \\
        & 90\% & $91.013_{\pm 0.038}$ & $62.399_{\pm 0.206}$ & $55.460_{\pm 0.421}$ \\
        \midrule
        \multirow{3}{*}{FlexCare} 
        & 56\% & $91.637_{\pm 0.048}$ & $67.242_{\pm 0.281}$ & $60.198_{\pm 0.218}$ \\
        & 75\% & $91.544_{\pm 0.038}$ & $66.858_{\pm 0.289}$ & $60.134_{\pm 0.346}$ \\
        & 90\% & $91.518_{\pm 0.029}$ & $66.426_{\pm 0.317}$ & $56.656_{\pm 0.376}$ \\
        \midrule
        \multirow{3}{*}{Diffmv} 
        & 56\% & $91.464_{\pm 0.056}$ & $66.389_{\pm 0.312}$ & $58.124_{\pm 0.187}$ \\
        & 75\% & $91.443_{\pm 0.037}$ & $65.615_{\pm 0.319}$ & $57.202_{\pm 0.228}$ \\
        & 90\% & $91.063_{\pm 0.029}$ & $64.443_{\pm 0.288}$ & $57.056_{\pm 0.205}$ \\
        \midrule
        \multirow{3}{*}{MUSE} 
        & 56\% & $91.359_{\pm 0.057}$ & $65.881_{\pm 0.328}$ & $57.224_{\pm 0.277}$ \\
        & 75\% & $91.207_{\pm 0.032}$ & $65.491_{\pm 0.579}$ & $55.936_{\pm 0.423}$ \\
        & 90\% & $91.064_{\pm 0.028}$ & $65.424_{\pm 0.413}$ & $52.392_{\pm 0.318}$ \\
        \midrule
        \multirow{3}{*}{MoSARe} 
        & 56\% & $91.565_{\pm 0.081}$ & $65.568_{\pm 0.336}$ & $59.566_{\pm 0.289}$ \\
        & 75\% & $91.311_{\pm 0.040}$ & $64.991_{\pm 0.207}$ & $58.850_{\pm 0.331}$ \\
        & 90\% & $90.486_{\pm 0.028}$ & $64.498_{\pm 0.172}$ & $58.252_{\pm 0.195}$ \\
        \midrule
        \multirow{3}{*}{\textbf{HP}} 
        & 56\% & $\mathbf{92.138_{\pm 0.052}}$ & $\mathbf{68.567_{\pm 0.381}}$ & $\mathbf{63.367_{\pm 0.356}}$ \\
        & 75\% & $\mathbf{91.856_{\pm 0.036}}$ & $\mathbf{68.061_{\pm 0.310}}$ & $\mathbf{63.277_{\pm 0.302}}$ \\
        & 90\% & $\mathbf{91.808_{\pm 0.027}}$ & $\mathbf{67.333_{\pm 0.196}}$ & $\mathbf{62.248_{\pm 0.333}}$ \\
        \bottomrule
    \end{tabular}
    }
\end{table}


\begin{table}[t]
    \centering
    \caption{Performance comparison under the \textit{Only Modality Missing} setting on MIMIC-III dataset.}
    \label{tab:only_modality_missing_details}
    \resizebox{0.75\linewidth}{!}{
    \begin{tabular}{lccc}
        \toprule
        Method & AUROC & AUPRC & F1 \\
        \midrule
        MIPM & $92.085_{\pm 0.089}$ & $69.448_{\pm 0.122}$ & $62.840_{\pm 0.155}$ \\
        RedCore & $92.168_{\pm 0.153}$ & $68.148_{\pm 0.455}$ & $60.632_{\pm 0.722}$ \\
        FlexCare & $92.113_{\pm 0.098}$ & \uline{69.943}$_{\pm 0.137}$ & \uline{62.410}$_{\pm 0.310}$ \\
        Diffmv & $91.821_{\pm 0.063}$ & $68.674_{\pm 0.289}$ & $59.633_{\pm 0.435}$ \\
        MUSE & $92.178_{\pm 0.048}$ & $69.568_{\pm 0.219}$ & $62.352_{\pm 0.336}$ \\
        MoSARe & \uline{92.270}$_{\pm 0.055}$ & $68.032_{\pm 0.177}$ & $60.765_{\pm 0.240}$ \\
        \textbf{HP} & $\mathbf{92.557_{\pm 0.039}}$ & $\mathbf{70.015_{\pm 0.126}}$ & $\mathbf{64.133_{\pm 0.371}}$ \\
        \bottomrule
    \end{tabular}
    }
\end{table}


\begin{table}[t]
    \centering
    \caption{Performance comparison under the \textit{Only Label Missing} setting on MIMIC-III dataset.}
    \label{tab:only_label_missing_details}
    \resizebox{0.75\linewidth}{!}{
    \begin{tabular}{lccc}
        \toprule
        Method & AUROC & AUPRC & F1 \\
        \midrule
        MIPM & $82.821_{\pm 0.095}$ & $42.707_{\pm 0.375}$ & $40.237_{\pm 1.211}$ \\
        PRIME & $82.971_{\pm 0.139}$ & $42.698_{\pm 0.716}$ & $41.282_{\pm 0.789}$ \\
        MEDHMP & $85.106_{\pm 0.101}$ & $42.234_{\pm 1.353}$ & $40.538_{\pm 0.935}$ \\
        VecoCare & $82.167_{\pm 0.117}$ & $42.043_{\pm 0.648}$ & \uline{43.088}$_{\pm 0.919}$ \\
        MUSE & $80.942_{\pm 0.151}$ & $38.133_{\pm 0.285}$ & $38.565_{\pm 0.517}$ \\
        MoSARe & \uline{85.640}$_{\pm 0.218}$ & \uline{45.065}$_{\pm 0.503}$ & $39.021_{\pm 1.723}$ \\
        \textbf{HP} & $\mathbf{85.686_{\pm 0.152}}$ & $\mathbf{51.414_{\pm 0.515}}$ & $\mathbf{51.301_{\pm 0.650}}$ \\
        \bottomrule
    \end{tabular}
    }
\end{table}

\subsection{Detailed Robustness Results}
\label{app:robustness}

This section reports the detailed numerical results for the robustness analysis in the main text. \tabref{tab:label_robustness} and \tabref{tab:modality_robustness} present performance under varying label missing rates \{25\%, 50\%, 75\%, 90\%\} and modality missing rates \{56\%, 75\%, 90\%\}, respectively. \tabref{tab:only_modality_missing_details} and \tabref{tab:only_label_missing_details} further report results under the \textit{Only Modality Missing} and \textit{Only Label Missing} settings. In both settings, temporal irregularity is retained as an inherent property of raw EHRs.
Overall, HP remains consistently strong across diverse and severe incompleteness settings.

\subsection{Detailed Efficiency and Performance Analysis}
\label{app:efficiency_details}

This section reports the detailed results for the efficiency analysis in the main text. \tabref{tab:efficiency_full} presents predictive performance and inference latency for all compared methods and HP variants. All latency results are measured with a batch size of 32, and ``Time (ms)'' denotes the average per-sample inference latency.

\begin{table}[t]
    \centering
    \caption{Detailed comparison of model performance and inference efficiency. Inference time is reported in ms/sample.}
    \label{tab:efficiency_full}
    \renewcommand{\arraystretch}{1.1} 
    \setlength{\tabcolsep}{6pt} 
    \resizebox{0.8\linewidth}{!}{
    \begin{tabular}{lcccc}
        \toprule
        \textbf{Method} & \textbf{AUROC } & \textbf{AUPRC} & \textbf{F1} & \textbf{Time (ms)} \\
        \midrule
        MIPM & 91.621 & 67.197 & 60.239 & 29.39 \\
        PRIME & 91.537 & 66.625 & 59.518 & 29.83 \\
        MEDHMP & 90.091 & 63.842 & 55.423 & 14.41 \\
        VecoCare & 90.234 & 61.692 & 55.522 & 15.70 \\
        HEART & 90.222 & 62.889 & 56.893 & 17.69 \\
        MulT-EHR & 90.296 & 62.957 & 56.245 & 14.66 \\
        M3Care & 90.357 & 63.433 & 57.201 & 19.51 \\
        UMM & 88.359 & 59.492 & 54.434 & 18.85 \\
        DrFuse & 89.819 & 62.713 & 57.359 & 16.38 \\
        RedCore & 91.710 & 67.169 & 60.316 & 14.65 \\
        FlexCare & 91.637 & 67.242 & 60.198 & 18.37 \\
        Diffmv & 91.464 & 66.389 & 58.124 & 26.03 \\
        MUSE & 91.359 & 65.881 & 57.224 & 19.71 \\
        MoSARe & 91.565 & 65.568 & 59.566 & 17.84 \\
        \midrule
        HP \#1-4 \textbar\ 1-4 & 92.339 & 68.898 & 63.289 & 24.57 \\
        HP \#1-4 \textbar\ 4-12 & 92.138 & 68.567 & 63.367 & 17.78 \\
        HP \#2-12 \textbar\ 4-12 & 92.007 & 68.494 & 61.546 & 14.24 \\
        HP \#1-4 \textbar\ 12-24 & 92.048 & 67.955 & 60.922 & 13.92 \\
        \bottomrule
    \end{tabular}
    }
\end{table}

\subsection{Detailed Ablation Analysis}
\label{app:ablation_details}

This section provides additional ablation results.

\textbf{i) Cross-domain Interaction.}
We ablate the Cross-Modality and Cross-Case LRRL modules. As shown in \tabref{tab:ablation_appendix}, removing either module degrades performance, indicating that both cross-modal fusion and cross-case interaction contribute to more complete patient modeling and more robust modality recovery.

\textbf{ii) Low-rank Calculation Components.}
We further ablate the \textit{coupled term} and the \textit{unary term} in Eq.~\ref{eq:low_rank}, and the results are given in \tabref{tab:ablation_appendix}. Removing either term reduces performance, showing that both components are necessary: the \textit{coupled term} ($\sum Z_{ij}^{(\gamma)}$) captures high-order cross-dimensional dependencies, while the \textit{unary term} ($\mathbf{w}_{*}^\top \bm{r}_{ij}^{*}$) preserves first-order linear effects. Their combination enables a more complete characterization of clinical point relations.

\begin{table}[t]
    \centering
    \caption{Additional ablation study on Cross-domain Interaction and Low-rank calculation details (MIMIC-III).}
    \label{tab:ablation_appendix}
    \resizebox{0.9\linewidth}{!}{
    \begin{tabular}{lccc}
        \toprule
        \textbf{Variant} & \textbf{AUROC (\%)} & \textbf{AUPRC (\%)} & \textbf{F1 (\%)} \\
        \midrule
        \multicolumn{4}{l}{\textit{\textbf{Cross-domain Interaction}}} \\
        w/o Cross-Modality LRRL & $92.068_{\pm0.132}$ & $68.358_{\pm0.380}$ & $62.547_{\pm0.226}$ \\
        w/o Cross-Case LRRL & $91.707_{\pm0.063}$ & $67.501_{\pm0.237}$ & $62.368_{\pm0.302}$ \\
        \midrule
        \multicolumn{4}{l}{\textit{\textbf{Low-rank Calculation Details}}} \\
        w/o coupled term & $91.728_{\pm0.050}$ & $68.302_{\pm0.284}$ & $63.066_{\pm0.403}$ \\
        w/o unary term & $91.758_{\pm0.045}$ & $68.203_{\pm0.397}$ & $61.925_{\pm0.415}$ \\
        \midrule
        \textbf{HP (Full)} & $\bm{92.138}_{\pm 0.052}$ & $\bm{68.567}_{\pm 0.381}$ & $\bm{63.367}_{\pm 0.356}$ \\
        \bottomrule
    \end{tabular}
    }
\end{table}

\begin{table}[t]
    \centering
    \caption{Ablation study on entropy-based inference under different missing settings (MIMIC-III).}
    \label{tab:ablation_entropy}
    \resizebox{0.9\linewidth}{!}{
    \begin{tabular}{lccc}
        \toprule
        \textbf{Variant} & \textbf{AUROC (\%)} & \textbf{AUPRC (\%)} & \textbf{F1 (\%)} \\
        \midrule
        \multicolumn{4}{l}{\textbf{w/o Entropy}} \\
        Main Experiment
        & $91.888_{\pm0.038}$ & $68.493_{\pm0.592}$ & $\bm{63.538}_{\pm0.512}$ \\
        75\% Label Missing
        & $91.198_{\pm0.076}$ & $65.792_{\pm0.601}$ & $60.146_{\pm0.365}$ \\
        90\% Label Missing
        & $89.100_{\pm0.068}$ & $61.339_{\pm0.388}$ & $55.013_{\pm0.319}$ \\
        75\% Modality Missing
        & $91.751_{\pm0.036}$ & $67.191_{\pm0.290}$ & $62.850_{\pm0.277}$ \\
        90\% Modality Missing
        & $91.660_{\pm0.029}$ & $66.654_{\pm0.275}$ & $61.263_{\pm0.389}$ \\
        \midrule
        \multicolumn{4}{l}{\textbf{HP (Full)}} \\
        Main Experiment
        & $\bm{92.138}_{\pm0.052}$ & $\bm{68.567}_{\pm0.381}$ & $63.367_{\pm0.356}$ \\
        75\% Label Missing
        & $\bm{91.223}_{\pm0.103}$ & $\bm{66.078}_{\pm0.226}$ & $\bm{60.659}_{\pm0.398}$ \\
        90\% Label Missing
        & $\bm{90.176}_{\pm0.167}$ & $\bm{63.543}_{\pm0.414}$ & $\bm{58.489}_{\pm0.358}$ \\
        75\% Modality Missing
        & $\bm{91.856}_{\pm0.036}$ & $\bm{68.061}_{\pm0.310}$ & $\bm{63.277}_{\pm0.302}$ \\
        90\% Modality Missing
        & $\bm{91.808}_{\pm0.027}$ & $\bm{67.333}_{\pm0.196}$ & $\bm{62.248}_{\pm0.333}$ \\
        \bottomrule
    \end{tabular}
    }
\end{table}

\textbf{iii) Adaptive Entropy-based Inference.}
During inference, we apply an Adaptive Entropy-based Inference strategy for robustness. Specifically, the trained prediction heads attached to the 2nd-, 3rd-, and 5th-layer LRRL outputs respectively produce logits from unimodal, within-case fused, and cross-case fused representations. We compute the entropy of these logits and select the prediction with the lowest entropy as the final output.

As shown in \tabref{tab:ablation_entropy}, compared with directly using the final-layer output, this strategy consistently improves performance, with larger gains under higher missing rates. A likely reason is that under severe incompleteness, multimodal fusion is not always superior to relying on a single reliable modality, since recovery and fusion may propagate noise from missing or weak modalities. Entropy-based selection mitigates this issue by adaptively choosing the most confident representation level.

\subsection{Parameter Sensitivity Analysis}
\label{app:sensitivity}

We evaluate the sensitivity of the key hyperparameters in HP, including the rank $R$ in Low-rank Relational Attention, the sampling intervals in the Low-Rank Relational Sampled Layer, and the loss weights $\lambda_a$ and $\lambda_r$. All hyperparameters are selected from predefined candidate sets based on validation performance, and the detailed results are reported in \tabref{tab:sensitivity_rank}--\tabref{tab:sensitivity_lambda_r}.

 \textbf{1) Rank $R$.}
We vary $R \in \{4, 8, 16\}$, and the results are shown in \tabref{tab:sensitivity_rank}. Based on the overall performance, we select $R=8$ for both datasets.

 \textbf{2) Sampling Intervals.}
We evaluate different sampling interval settings for each dataset, and the results are reported in \tabref{tab:sensitivity_sampling}. For MIMIC-III, we select \textbf{1-4 \textbar\ 4-12}; for MIMIC-IV, we select \textbf{1-4 \textbar\ - \textbar\ 12-12}.

 \textbf{3) Loss Weights ($\lambda_a$ and $\lambda_r$).}
We further study the effects of the loss weights for Fine-grained Alignment and Fine-grained Reconstruction. The results are reported in \tabref{tab:sensitivity_lambda_a} and \tabref{tab:sensitivity_lambda_r}. According to the overall performance, we use $\lambda_a=0.002$ and $\lambda_r=10$ for MIMIC-III, and $\lambda_a=0.0001$ and $\lambda_r=5$ for MIMIC-IV.


\begin{table}[h]
    \centering
    \caption{Parameter sensitivity analysis of the Rank ($R$) in Low-rank Relational Attention.}
    \label{tab:sensitivity_rank}
    \resizebox{0.75\linewidth}{!}{
    \begin{tabular}{lccc}
        \toprule
        \textbf{Variant} & \textbf{AUROC (\%)} & \textbf{AUPRC (\%)} & \textbf{F1 (\%)} \\
        \midrule
        \multicolumn{4}{l}{\textit{\textbf{MIMIC-III}}} \\
        $R=4$ & $91.528_{\pm0.083}$ & $67.594_{\pm0.369}$ & $62.296_{\pm0.403}$ \\
        $R=8$ & $\bm{92.138}_{\pm0.052}$ & $\bm{68.567}_{\pm0.381}$ & $\bm{63.367}_{\pm0.356}$ \\
        $R=16$ & $91.938_{\pm0.030}$ & $68.321_{\pm0.205}$ & $63.328_{\pm0.288}$ \\
        \midrule
        \multicolumn{4}{l}{\textit{\textbf{MIMIC-IV}}} \\
        $R=4$ & $97.936_{\pm0.032}$ & $92.675_{\pm0.125}$ & $86.822_{\pm0.261}$ \\
        $R=8$ & $\bm{97.980}_{\pm0.033}$ & $\bm{93.207}_{\pm0.103}$ & $\bm{87.203}_{\pm0.209}$ \\
        $R=16$ & $97.988_{\pm0.045}$ & $92.988_{\pm0.357}$ & $87.166_{\pm0.356}$ \\
        \bottomrule
    \end{tabular}
    }
\end{table}

\begin{table}[!ht]
    \centering
    \caption{Parameter sensitivity analysis of Sampling Intervals. }
    \label{tab:sensitivity_sampling}
    \resizebox{0.8\linewidth}{!}{
    \begin{tabular}{lccc}
        \toprule
        \textbf{Sampling Interval} & \textbf{AUROC (\%)} & \textbf{AUPRC (\%)} & \textbf{F1 (\%)} \\
        \midrule
        \multicolumn{4}{l}{\textit{\textbf{MIMIC-III}} ($m^{(1)}$ \textbar\ $m^{(2)}$)} \\
        1-4 \textbar\ 1-4 & $92.339_{\pm0.107}$ & $68.898_{\pm0.293}$ & $63.289_{\pm0.437}$ \\
        \textbf{1-4 \textbar\ 4-12} & $\bm{92.138}_{\pm0.052}$ & $\bm{68.567}_{\pm0.381}$ & $\bm{63.367}_{\pm0.356}$ \\
        1-4 \textbar\ 12-24 & $92.048_{\pm0.062}$ & $67.955_{\pm0.425}$ & $60.922_{\pm0.297}$ \\
        2-12 \textbar\ 4-12 & $92.007_{\pm0.053}$ & $68.494_{\pm0.318}$ & $61.546_{\pm0.441}$ \\
        2-12 \textbar\ 12-24 & $91.776_{\pm0.159}$ & $68.177_{\pm0.327}$ & $61.328_{\pm0.408}$ \\
        \midrule
        \multicolumn{4}{l}{\textit{\textbf{MIMIC-IV}} ($m^{(1)}$ \textbar\ $m^{(2)}$ \textbar\ $m^{(3)}$)} \\
        \textbf{1-4 \textbar\ - \textbar\ 12-12} & $\bm{97.980}_{\pm0.033}$ & $\bm{93.207}_{\pm0.103}$ & $\bm{87.203}_{\pm0.209}$ \\
        1-4 \textbar\ - \textbar\ 12-24 & $97.971_{\pm0.039}$ & $92.887_{\pm0.158}$ & $87.085_{\pm0.225}$ \\
        1-4 \textbar\ - \textbar\ 12-48 & $97.929_{\pm0.042}$ & $92.736_{\pm0.218}$ & $86.860_{\pm0.235}$ \\
        \bottomrule
    \end{tabular}
    }
\end{table}

\begin{table}[!ht]
    \centering
    \caption{Sensitivity analysis of $\lambda_a$.}
    \label{tab:sensitivity_lambda_a}
    \resizebox{0.8\linewidth}{!}{
    \begin{tabular}{lccc}
        \toprule
        \textbf{$\lambda_a$} & \textbf{AUROC (\%)} & \textbf{AUPRC (\%)} & \textbf{F1 (\%)} \\
        \midrule
        \multicolumn{4}{l}{\textit{\textbf{MIMIC-III}}} \\
        0.02 & $92.181_{\pm0.056}$ & $68.283_{\pm0.277}$ & $62.161_{\pm0.305}$ \\
        \textbf{0.002} & $\bm{92.138}_{\pm0.052}$ & $\bm{68.567}_{\pm0.381}$ & $\bm{63.367}_{\pm0.356}$ \\
        0.001 & $92.052_{\pm0.043}$ & $68.760_{\pm0.360}$ & $63.040_{\pm0.299}$ \\
        0.0002 & $91.762_{\pm0.108}$ & $67.843_{\pm0.405}$ & $61.506_{\pm0.401}$ \\
        0 (w/o FGA) & $91.926_{\pm0.031}$ & $67.546_{\pm0.258}$ & $61.427_{\pm0.290}$ \\
        \midrule
        \multicolumn{4}{l}{\textit{\textbf{MIMIC-IV}}} \\
        0.01 & $97.852_{\pm0.035}$ & $93.067_{\pm0.155}$ & $87.122_{\pm0.225}$ \\
        0.001 & $97.883_{\pm0.028}$ & $93.187_{\pm0.120}$ & $87.220_{\pm0.185}$ \\
        \textbf{0.0001} & $\bm{97.980}_{\pm0.033}$ & $\bm{93.207}_{\pm0.103}$ & $\bm{87.203}_{\pm0.209}$ \\
        0.00001 & $97.925_{\pm0.025}$ & $92.989_{\pm0.098}$ & $86.890_{\pm0.215}$ \\
        0.000001 & $97.870_{\pm0.068}$ & $92.786_{\pm0.177}$ & $86.928_{\pm0.305}$ \\
        0 (w/o FGA) & $97.931_{\pm0.037}$ & $92.874_{\pm0.085}$ & $86.851_{\pm0.290}$ \\
        \bottomrule
    \end{tabular}
    }
\end{table}

\begin{table}[!ht]
    \centering
    \caption{Sensitivity analysis of $\lambda_r$.}
    \label{tab:sensitivity_lambda_r}
    \resizebox{0.8\linewidth}{!}{
    \begin{tabular}{lccc}
        \toprule
        \textbf{$\lambda_r$} & \textbf{AUROC (\%)} & \textbf{AUPRC (\%)} & \textbf{F1 (\%)} \\
        \midrule
        \multicolumn{4}{l}{\textit{\textbf{MIMIC-III}}} \\
        0 (w/o FGR) & $91.823_{\pm0.055}$ & $67.784_{\pm0.276}$ & $61.593_{\pm0.317}$ \\
        1 & $91.551_{\pm0.048}$ & $67.232_{\pm0.329}$ & $62.127_{\pm0.303}$ \\
        \textbf{10} & $\bm{92.138}_{\pm0.052}$ & $\bm{68.567}_{\pm0.381}$ & $\bm{63.367}_{\pm0.356}$ \\
        100 & $92.113_{\pm0.105}$ & $67.889_{\pm0.398}$ & $63.317_{\pm0.364}$ \\
        \midrule
        \multicolumn{4}{l}{\textit{\textbf{MIMIC-IV}}} \\
        1 & $97.922_{\pm0.058}$ & $92.845_{\pm0.153}$ & $85.829_{\pm0.597}$ \\
        \textbf{5} & $\bm{97.980}_{\pm0.033}$ & $\bm{93.207}_{\pm0.103}$ & $\bm{87.203}_{\pm0.209}$ \\
        10 & $97.917_{\pm0.039}$ & $92.661_{\pm0.119}$ & $86.698_{\pm0.265}$ \\
        \bottomrule
    \end{tabular}
    }
\end{table}

\FloatBarrier
\section*{CRediT authorship contribution statement}
\textbf{Bohao Li:} Writing -- review \& editing, Writing -- original draft, Visualization, Validation, Software, Project administration, Methodology, Formal analysis, Data curation, Conceptualization.
\textbf{Tao Zou:} Writing -- review \& editing.
\textbf{Junchen Ye:} Writing -- review \& editing, Project administration, Conceptualization.
\textbf{Yan Gong:} Writing -- review \& editing.
\textbf{Bowen Du:} Project administration, Investigation, Funding acquisition.

\section*{Declaration of competing interest}
The authors declare that they have no known competing financial interests or personal relationships that could have appeared to influence the work reported in this paper.

\section*{Data availability}
The data used in this study are available through PhysioNet: MIMIC-III (\url{https://physionet.org/content/mimiciii/}), MIMIC-IV (\url{https://physionet.org/content/mimiciv/}), MIMIC-IV-Note (\url{https://physionet.org/content/mimic-iv-note/}), and MIMIC-CXR (\url{https://physionet.org/content/mimic-cxr/}).
Access requires PhysioNet credentialing, completion of the required training, and acceptance of the applicable data use agreements. The authors cannot redistribute the underlying patient-level data under these agreements.

\section*{Declaration of generative AI and AI-assisted technologies in the manuscript preparation process}
During the preparation of this work, the authors used AI agent to assist with language editing, and LaTeX formatting. After using this tool, the authors reviewed and edited the content as needed and take full responsibility for the content of the published article.

\bibliographystyle{elsarticle-num}
\bibliography{reference}
\end{document}